\pdfoutput=1
\documentclass[11pt]{article}

\usepackage[final]{acl}

\usepackage{times}
\usepackage{latexsym}

\usepackage[T1]{fontenc}
\usepackage[utf8]{inputenc}

\usepackage{microtype}

\usepackage{inconsolata}

\usepackage{graphicx}

\usepackage{mathtools}
\usepackage{booktabs}

\usepackage{hyperref}

\title{What We Talk About When We Talk About LMs:\\Implicit Paradigm Shifts and the \textit{Ship of Language Models}}

\author{Shengqi Zhu \\
  Cornell University \\
  \texttt{sz595@cornell.edu} \\\And
  Jeffrey M. Rzeszotarski \\
  Cornell University \\
  \texttt{jeffrz@cornell.edu}}

\begin{document}
\maketitle
\begin{abstract}
The term Language Models (LMs) as a time-specific collection of models of interest is constantly reinvented, with its referents updated much like the \textit{Ship of Theseus} replaces its parts but remains the same ship in essence.
In this paper, we investigate this \textit{Ship of Language Models} problem, wherein scientific evolution takes the form of continuous, implicit retrofits of key \textit{existing} terms. We seek to initiate a novel perspective of scientific progress, in addition to the more well-studied emergence of \textit{new} terms.
To this end, we construct the data infrastructure based on recent NLP publications.
Then, we perform a series of text-based analyses toward a detailed, quantitative understanding of the use of Language Models as a term of art. 
Our work highlights how systems and theories influence each other in scientific discourse, and we call for attention to the transformation of this Ship to which we all are contributing.\footnote{The data and code of this work is available at \url{https://github.com/CurlyZhu/Ship_of_LMs}.}
\end{abstract}

\section{Introduction}

Scientific publications expand exponentially, with the size of literature doubling every $\sim$17 years~\cite{fortunato2018science,bornmann2021growth}.
The field of CL/NLP is no exception;
in fact, the doubling only took 5 years: As of 2023, the number of papers documented in the ACL Anthology is twice as much as the total by 2018~\cite{bollmann2023two,zhao2023survey}.
With the explosion of new publications, it is imperative but also increasingly challenging to sort out the major contexts, progress, and future directions of the field.

Researchers have sought to identify emergent key terms and factors that led to disruptive shifts of paradigms (\citealp{uban-etal-2021-studying,pramanick-etal-2023-diachronic,kuhn2014inheritance}).
In this period of flux, however, an ever-evolving field like ours calls for deeper analysis beyond \textit{identifying} these elements, in order to understand various \textit{quantitative} questions regarding these rapid and disruptive shifts. 
For instance, \textit{to what extent} is the field transformed by any certain model, like ChatGPT~\cite{chatgpt-openai}?
How does the popularity of the latest GPT models compare with, say, that of BERT~\cite{devlin-etal-2019-bert} in 2020?
From there, we might even be curious about bigger questions like, ``How unprecedented really is ChatGPT?'', where our empirical guesses can diverge drastically without sufficient quantitative evidence.
While readers of this paper likely come in with a tacit understanding of the ebb and flow of the field, it is hard to nail down such factors that keep changing in publications.

More fundamentally, the narrative describing scientific progress as the emergence of new elements does not cover the more \textit{implicit} paradigm shifts, which features the \textit{evolution} instead of \textit{invention} of terms.
The (forms of) key terms may continue to be broadly used, but are gradually overwritten with new meanings in new contexts.
\textit{Language Models} (LMs), as a term of art, refers to no single, static thing.
It is used referentially to index a collection of models deemed relevant and representative at the time or in the context of a paper. 
As this is ever-changing, we are faced with a \textit{Ship of Theseus} scenario \cite{plutarch-ship-of-theseus}, wherein the same terminology is essentially re-invented and its referents are perhaps entirely replaced.
%\begin{quote}
    %The ship wherein Theseus and the youth of Athens returned had thirty oars, and was preserved by the Athenians down even to the time of Demetrius Phalereus, for they took away the old planks as they decayed, putting in new and stronger timber in their place, insomuch that this ship became a standing example among the philosophers, for the logical question of things that grow; one side holding that the ship remained the same, and the other contending that it was not the same. (Plutarch, “Life of Theseus”)
%\end{quote}
As such, a subtle gap between the durable collective \textit{term} of ``LMs'' and the time-specific \textit{referent} models of the moment is widening as the field progresses, threatening its stability and accessibility to new researchers.
%We might wonder if a finding on LMs from a while ago remains effective, or what model selections sound representative of and generalizable across LMs.
These issues call for new analyses of the subtle transformations that result from these paradigm shifts.

In this paper, we seek a quantitative description of a field's continuous evolution.
More specifically, we inquire into the \textit{Ship of LMs} paradox, i.e., the aforementioned reconstruction of the term \textit{Language Models}. We decipher this evergreen term's rapidly changing referents, contexts, and usages across time.
We develop a semi-automatic, generalizable framework to extract and organize two closely related sets of keywords: (1) mentions of the collective LM concept, and (2) specific model names, and construct a dataset of 7,650 papers from the 10 most recent major NLP conferences.

We focus on several questions concerning how we as a field talk about LMs: How often do we talk about LMs, and how confidently? (\S\ref{subsec:surging-passion}) Which models, and what is special about these components? (\S\ref{subsec:which-models}) Moreover, how do the referents of LM vary across papers? (\S\ref{subsec:diff_among_groups})
Finally, we conclude the findings and future perspectives in \S\ref{sec:discussion}.

Our work highlights the astonishing extent of change subtly encoded in the seemingly unchanged overarching terms.
We hope the \textit{Ship of LMs} can serve as a new perspective to understand the field's progress, and that our methodology can serve as an entry point for finer-grained measurements of subtle changes in rapidly growing scientific fields.

\section{Related Work}
\paragraph{Diachronic Analysis of the Progress in NLP} 
Various studies review the history of NLP conferences and the ACL Anthology~\cite{hall-etal-2008-studying,anderson-etal-2012-towards,bollmann2023two}, as well as the community that contributed to the field's trajectory~\cite{abdalla-etal-2023-elephant,movva2023large}. 
Other works identified the field's transition points and themes: \citet{hou-etal-2019-identification} proposes an automatic framework to extract key entities (tasks, dataset, etc.); \citet{uban-etal-2021-studying} explores a similar goal via topic modeling; and \citet{pramanick-etal-2023-diachronic} further identified such entities that causally shaped the field's important stages.
More recently, there has also been a specific focus on the changes brought by LLMs~\cite{min2023recent,fan2023bibliometric,zhao2023survey} and the impact on the related communities~\cite{saphra2023first,liang2024monitoring}.
Aside from text-based analysis, interviews and surveys (\citealp{gururaja2023build, michael-etal-2023-nlp}, etc.) have also provided valuable qualitative insights for the disruptive shifts.

\paragraph{Paradigm Shifts and Scientific Trends} have also been core topics in the broader Science of Science field~\cite{fortunato2018science} beyond CL/NLP.
The existing literature mostly centers on the emergence of new, trending ideas as well as their dynamics across the author networks.
For instance, \citet{kuhn2014inheritance} identified text snippets that are largely cited by future works, coined \textit{scientific memes}, on citation graphs;
\citet{cheng2023new} explored the diffusion process of new ideas under various social factors;
and \citet{chu2021slowed} measured the relation between the speed of producing new ideas and the size of a field.
Citation/Author networks have also been introduced by recent works~\cite{mohammad-2020-examining,wahle-etal-2023-cite} as a method for the more specific background of the NLP field.

Our work complements these ongoing threads.
As discussed, we raise a novel scenario about the transitions within a \textit{lasting} concept (\textit{Ship of LMs}), which to our knowledge has not been explored.
We examine the use of such terms as LMs, providing \textit{quantitative} interpretations of how (and how much) our beliefs and common grounds have evolved.
In some sense, our work can also be seen as a meta-analysis of the various works studying certain elements (e.g. ``the era of LLM'', ``stages of Statistical Machine Translation'', ``ChatGPT's impact'', etc.) We integrate these valuable findings to highlight a new question about the procedures: how exactly did we forge these of key elements into practice and eventually into our norms of language?

\section{Methods}
\label{section:methods}
\subsection{Dataset Construction}
Following common practice in prior work \cite{mohammad-2020-examining,pramanick-etal-2023-diachronic}, we utilize the official ACL Anthology as our data source.
We collect papers accepted to the main Proceedings of three major NLP conferences (ACL, EMNLP, NAACL) held annually\footnote{NAACL is merged with ACL once every three years, and thus there is no NAACL conference data in 2020 or 2023.}.
We first interact with the API to fetch metadata (e.g., Anthology ID, title, and abstract).
Based on the index of a conference, we obtain the paper PDFs from the formatted Anthology URLs, and scan the text with the \texttt{pypdf}\footnote{\url{https://pypdf.readthedocs.io/}} tools. For post-processing, we remove excessive formatting (e.g. conference names in the footers) and identify section titles with regular expressions.

The resulting dataset contains in total 7,650 papers from 10 conferences sequentially from ACL 2020 to EMNLP 2023. 
Our analyses focus on this most recent 4-year window where the advances regarding LMs have been especially pronounced, while our methodologies can similarly extend to a broader range.

\paragraph{Default Setup} We extract the \textit{body text} by cutting off before the References section. This is marked as our \textit{default setup}, and experiments are based on the default unless otherwise noted.

\begin{figure*}[t!]
    \centering
    \includegraphics[width=0.98\textwidth]{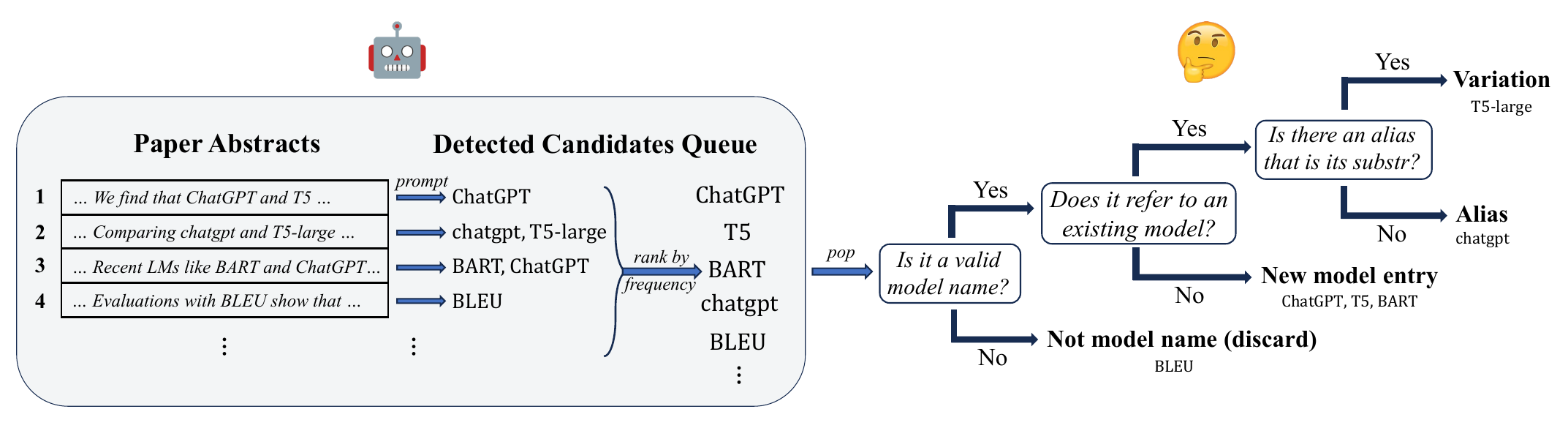}
    \caption{The full pipeline for constructing the model dictionaries (\S\ref{subsubsec:construct-m}). The LLM agent follows a formatted prompt (\S\ref{appendix:prompt}) to automatically identify potential model names. The extracted strings are merged and ranked by frequency to form the list of candidate names. Then, the authors manually validate whether it is a new entry, an alias, or other based on a fixed protocol: (1) whether the candidate is indeed a valid model name, (2) whether it refers to the same model as an existing entry, and (3) whether it's already covered by an alias of that entry.}
    \label{fig:workflow}
\end{figure*}

\subsection{Retrieving the Mentions of LMs}
To investigate the \textit{Ship of LMs} problem, we start by extracting and analyzing relevant keywords and entities, a common backbone method for analysis~\cite{hou-etal-2019-identification,pramanick-etal-2023-diachronic}.
For a sample paper to be related to LMs, the writing could utilize two types of mentions: (1) the collective concept of ``language models'', implying the context as a generalizable discussion, and (2) the names of specific models\footnote{We consider the broadest definition and scope of LMs; see Appendix~\ref{appendix:LM_definition} for a more detailed note.}, indicating what models are exactly considered in a limited scope.

Our goal is to maintain two keyword sets that correspond to the two types.
Thus, we can resolve the referents of the generic language model mentions to the specific models used, by locating, linking, and comparing the keywords from both sets.

\subsubsection{Notations}
As described, we seek to build two related collections of key entities, one marking the mentions of LMs as a general term, and the other marking specific model names.
The two are respectively denoted by $\mathcal{L}$ (from \textbf{L}M) and $\mathcal{M}$ (from \textbf{M}odels).

In practice, $\mathcal{L}$ converges to a small, well-recognized set of terms.
We define
\begin{center}
    $\mathcal{L}$ = \{language model, LLM, PLM\}
\end{center}
since ``language model'' is the substring of most of its subcategories, e.g., ``large language models'', ``Korean language models'', or ``language modeling'', and thus searching for ``language model''\footnote{Here we include other upper-case forms, e.g., ``Language Models''. This is implemented similar to \textit{Aliases} in \S\ref{subsubsec:construct-m}.} covers all such variations.
We also include the most common acronyms, ``LLM'' (Large Language Models) and ``PLM'' (Pre-trained Language Models). 
% Maybe add VLM in the future?
The construction of $\mathcal{M}$ is elaborated in \S\ref{subsubsec:construct-m}.

We use $m$ to represent a specific element from $\mathcal{M}$ (e.g., $m = $ BERT).
For a span of text, $s$, we have $\mathbf{M}(s)$ representing the subset of elements from $\mathcal{M}$ that indeed appear in $s$.\footnote{We can similarly define $l$ and $L(\cdot)$ based on $\mathcal{L}$. In practice, however, we don't further distinguish between different LM terms, given the limited size and high convergence of $\mathcal{L}$.} While $\mathbf{M}(\cdot)$ is a function of the input text by definition, we omit the input when $s$ is the entire body (default setup) for simplicity, and write the subset as $\mathbf{M}$. The omission applies similarly to the notations below.

To initiate our study on any individual paper and any model(s) of interest, we introduce a family of \textit{counting} functions. Given a model name $m$, we define $N_m(\cdot)$ as the count of how many times $m$ appears in the input text. The counting functions also apply to sets of model names: For a set of models $M = \{m_1, m_2, ..., m_k\}$, we have
\begin{equation*}
    \vspace*{-0.1cm}
    N_M = \sum_{i=1}^k N_{m_i}
    %\vspace*{-0.1cm}
\end{equation*}
Thus, $\mathbf{M}$ can now be formally defined as
\begin{equation*}
    \mathbf{M} = \{m \, | \, N_m > 0, \, \forall m \in \mathcal{M}\}
\end{equation*}
Additionally, given its importance, we mark the count of all model names in a paper as
\begin{equation*}
    N \coloneq N_\mathcal{M} = N_{\mathbf{M}}
\end{equation*}
Similarly for the other keyword set of general LM mentions, we denote the total count of all elements in $\mathcal{L}$ as $N^{\mathcal{L}}$. We mark $\mathcal{L}$ as superscript for an explicit distinction with the $N_m$ family. The total counts $N^{\mathcal{L}}$, $N$, and the $N_m$ family serve as essential cornerstones of our approach since they are direct indicators of how LMs are discussed and resolved. These patterns from independent works become the changing constituents of the Ship.

\subsubsection{Constructing \texorpdfstring{$\mathcal{M}$}{M} from the text}
\label{subsubsec:construct-m}
To construct a comprehensive dictionary of specific model names, we established a human-AI workflow to extract and register model names at scale.
We designed a detailed in-context prompt for a state-of-the-art LLM to detect model names from the title and abstract of papers.
All detected names from the full dataset are collected and ranked by frequencies as candidates $[\hat{m}_1, \hat{m}_2, ...]$.
Since the same type of model as referent can have various textual forms, we aim to maintain and distinguish two attributes (as lists) for a model $m$:
\begin{itemize}
    \item \textbf{Aliases}: Different text patterns that all refer to $m$; e.g., both ``chatgpt'' and ``ChatGPT'' are identified separately but point to the same thing, and we need to count them together.
    \item \textbf{Variations}: Refers to $m$, but is the extension of an existing alias (i.e. having an alias as substring). This usually suggests a specific variation of $m$, e.g., ``T5-3B'' when $m =$ ``T5''~\cite{raffel2020exploring}. Searching for ``T5'' in the text would have already included the mentions of ``T5-3B''.
\end{itemize}
To compose the final list, we manually validate and register the name candidates following a simple heuristic (Figure~\ref{fig:workflow}). When we encounter a new model $m$ not in $\mathcal{M}$, we add $m$ to $\mathcal{M}$ and initialize its alias list as $[m]$. Additional aliases of an entry $m_j$ are appended to its list. Variations of existing entries are recorded but not added as an alias, and candidates that are not the name of a model (e.g., BLEU) are discarded.

For each entry of $m$, we also manually retrieved the original paper or documentation to determine if there is an explicit dependency on another model. In all, $\mathcal{M}$ has a total of 103 model entries and 155 aliases. With the two keyword lists, $\mathcal{L}$ and $\mathcal{M}$, we are ready to examine how LMs are resolved and extract diachronic patterns.

Our dataset and code is available at \url{https://github.com/CurlyZhu/Ship_of_LMs}. We also provide more details of the implementation in the Appendix: the full list of models involved (\ref{appendix:full_model_entry}), the setup of the model name detector (\ref{appendix:LLM_setup}), and the LLM prompts (\ref{appendix:prompt}).

\section{Experiments and Findings}
\label{section:exp_and_findings}

LMs have been steadily gaining more attention from the field. \citet{zhao2023survey} reports that papers containing the key phrase ``Language Model'' have increased from less than 400 pre-2019 to around 10,000 in 2023. We observe a similar trend with a finer-grained search in the NLP domain (Figure~\ref{fig:LM_surge}(a)). At ACL 2020, 35\% of the papers contain at least one LM mention from $\mathcal{L}$ (we refer to this portion as \textbf{LM-related papers}). Since then, this proportion has had a smooth, continuous growth of approximately $5\%$ (additive) per conference, hitting 84\% just three years later at EMNLP 2023.

\begin{table}[t!]
    \centering
    \setlength\tabcolsep{4.35pt}
    \renewcommand{\arraystretch}{1.35}
    \scalebox{0.65}{
        \begin{tabular}{@{}|l|c|c|c|c|c|c|c|c|c|c|@{}} \hline
        \multicolumn{1}{|c|}{\textbf{Year} \textit{(prefix)}} & \multicolumn{2}{c|}{\textbf{20-}} & \multicolumn{3}{c|}{\textbf{21-}} & \multicolumn{3}{c|}{\textbf{22-}} & \multicolumn{2}{c|}{\textbf{23-}}   \\ \hline %\midrule
        \multicolumn{1}{|c|}{\textbf{Venue} \textit{(suffix)}}      & \textbf{A} & \textbf{E} & \textbf{N} & \textbf{A} & \textbf{E} & \textbf{A} & \textbf{N} & \textbf{E} & \textbf{A} & \textbf{E} \\ \hline %\midrule
        Total      & 778  & 750 & 477 & 709 & 846 & 700 & 442 & 827 & 1074 & 1047 \\ \hline
        LM-related & 273  & 337 & 239 & 333 & 454 & 407 & 282 & 571 & 794  & 877  \\ \hline
        \end{tabular}
    }
    \caption{Total numbers of conference papers in this work after pre-processing. We denote the conferences henceforth with their first letters (\textbf{A}CL, \textbf{E}MNLP, and \textbf{N}AACL), and denote the years as prefixes for brevity. For instance, `21-E' represents EMNLP 2021.}
    \label{tab:basic_stats}
\end{table}

\subsection{Wind in the Sails: Surging Mentions, Speeding Conclusions}
\label{subsec:surging-passion}

We begin by querying a fundamental aspect of LMs' increasing popularity: Has our \textit{use} of the term \textit{LM} also evolved per se, apart from the background increase noted above? As one hypothesis, LMs' popularity might be attributed mainly to the increase of share. The types of work we do and the context of LMs may have not changed significantly -- it's just more authors working on the topic, more resources put into it, or other external factors.

We consider the average $N^\mathcal{L}$ of all papers at a certain conference, written as $\bar{N^\mathcal{L}}$. With the division of LM-related and non-related papers, $\bar{N^\mathcal{L}}$ can be calculated as the (weighted) average of the two groups. If the above assumption holds true, $\bar{N^\mathcal{L}}$ can be estimated as follows: On the one hand, non-LM-related papers always have $N^\mathcal{L} = 0$ by definition. On the other hand, within the LM-related group, the average of $N^\mathcal{L}$ would also be stable across conferences since the use of the term LMs remains generally unchanged as hypothesized. As both groups -- constituting the full set of papers at a conference when combined -- see no qualitative change in their within-group $N^\mathcal{L}$, the determining factor for the overall average $\bar{N^\mathcal{L}}$ is the ratio of the two groups’ size (as the weights in weighted average), i.e., the percentage of LM-related papers. Thus, we can draw an estimate from the scale
of the first data point ($\bar{N^\mathcal{L}}=4.29$ for $35\%$ LM-related at ACL 2020). For instance, 54\% of papers at EMNLP 2021 are LM-related, which is $1.54\times$ that of ACL 2020 (35\%). We can thus scale $\bar{N^\mathcal{L}}$ with the same ratio, $1.54 \times 4.29 \approx 6.56$, as the estimate for EMNLP 2021.

\begin{figure}[t!]
    \centering
    \includegraphics[width=0.94\columnwidth]{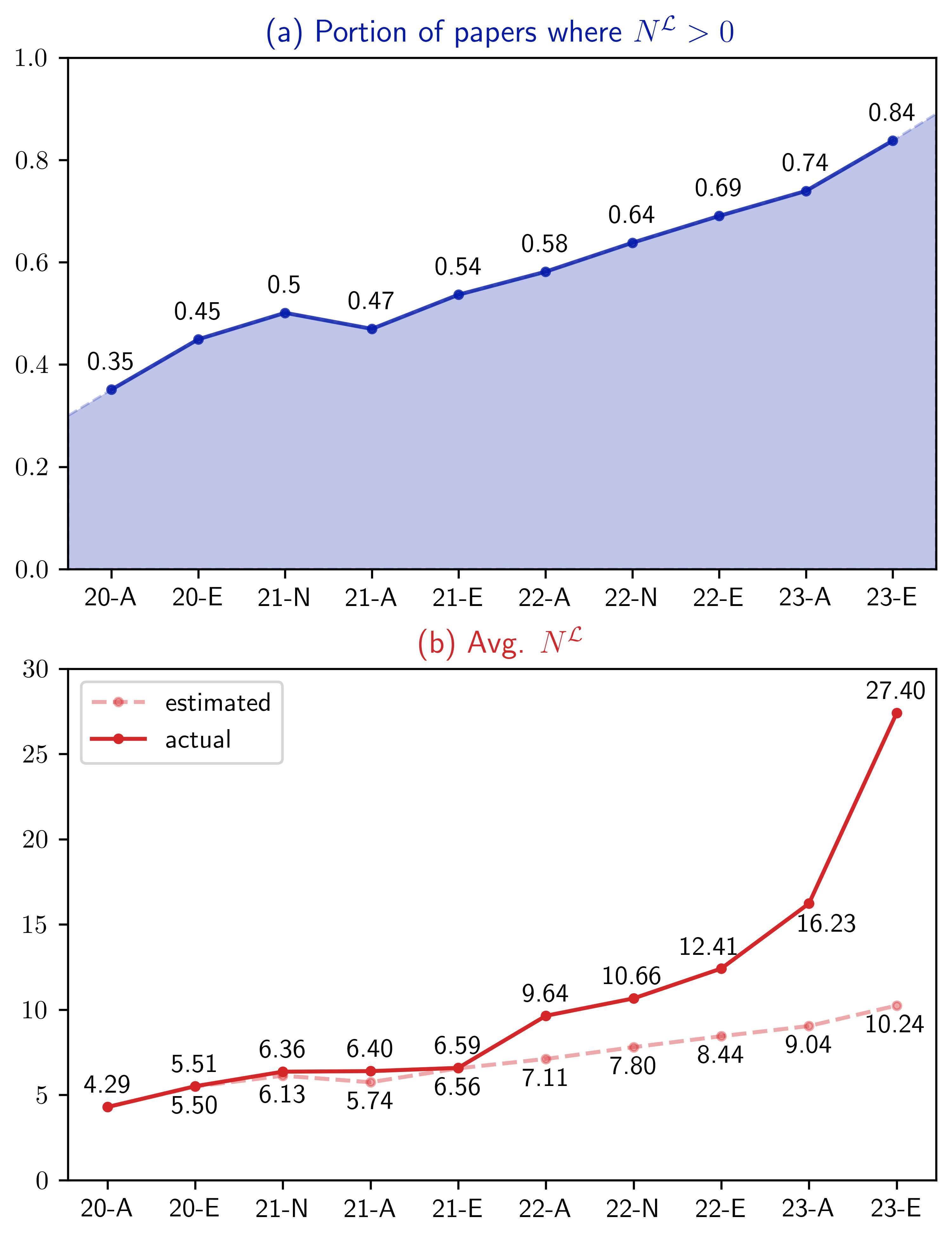}
    %\vspace*{-0.2cm}
    \caption{Increase of interest in LMs as a \textit{topic} (a) and as a \textit{term in use} (b). \textbf{(a)}: The proportion of papers containing keywords in $\mathcal{L}$ by years. \textbf{(b)}: The estimated value of $\bar{N^\mathcal{L}}$ based on the proportions (dashed line) compared with the actual $\bar{N^\mathcal{L}}$ (solid line).}
    \label{fig:LM_surge}
    %\vspace*{-0.5cm}
\end{figure}

Fig.~\ref{fig:LM_surge}(b) compares the actual $\bar{N^\mathcal{L}}$ and the value estimated in this way.
This turns out to be a surprisingly good fit for the first half of the data.
Within the first 5 conferences, the deviations between estimated and actual value are consistently less than 10\% and often close to 0. 
For this period, LMs gained more attention as a \textit{topic} in this period, but language describing this \textit{term} remained similar. Metaphorically, the composition of the Ship remains the same, but it has more wind in its sails.

However, we see a strong deviation from the estimated growth starting 2022. $\bar{N^\mathcal{L}}$ has since been on an exponential growth, eventually being 80\% higher than estimated at ACL 2023 and 168\% higher at EMNLP 2023 (where $\bar{N^\mathcal{L}}$ nearly doubled in just half a year).
In other words, the Ship is not just sailing better (more papers), but it is also undergoing reconstruction (referents of LM are changing).
The distinct patterns pre- and post-2022 despite a similar background increase highlight the necessity to study the Ship of LMs as a dynamic concept, as emergence of a term is not sufficient for mining the deeper nuances as such.

\begin{figure}[t!]
    \centering
    \includegraphics[width=0.82\columnwidth]{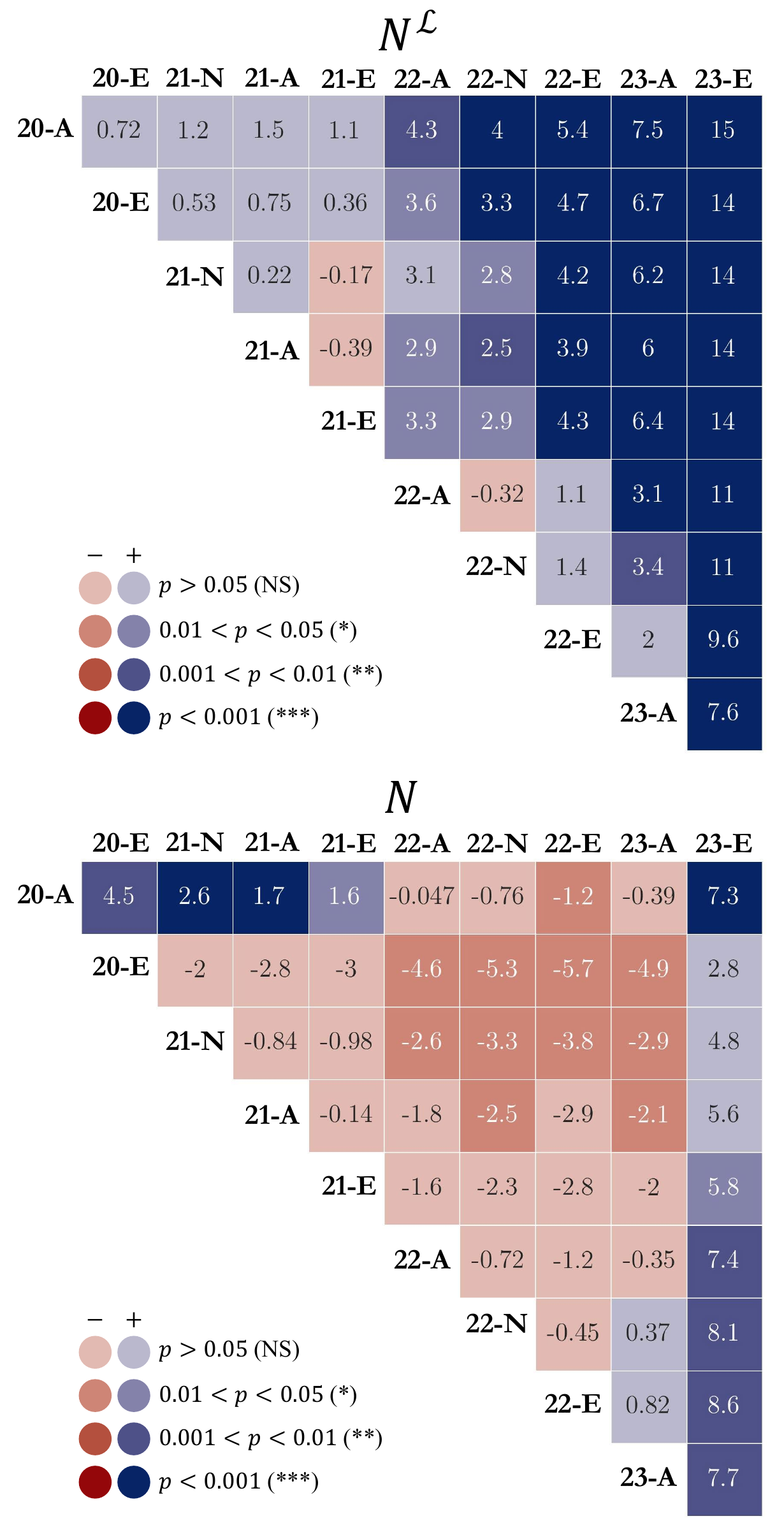}
    %\caption{KDE (Marginal Distribution) on the same scale}
    %\vspace*{-0.1cm}
    \caption{Pairwise comparisons of the distributions of $N^\mathcal{L}$ (upper) and $N$ (lower). Each grid corresponds to a pair of conferences indexed by its row and column, and depicts results from two analyses: a K-S test of whether the data of the pair are from different distributions (heatmap), and mean difference (digits and hue).}
    \label{fig:NL_vs_N}
    %\vspace*{-0.5cm}
\end{figure}

\paragraph{What about the \textit{actual} models we use?}
The super-linear increase of $N^{\mathcal{L}}$ demands investigation into its likely causes.
Authors might seek to cover more models in more detail, and their writing adapts to the strengthened claims, leading to the growth observed.
Alternatively, authors might be more eager to employ trending terms even without significantly stronger evidence or fit to their work.

\begin{figure*}[t!]
    \centering
    %\vspace*{-0.2cm}
    \includegraphics[width=\columnwidth]{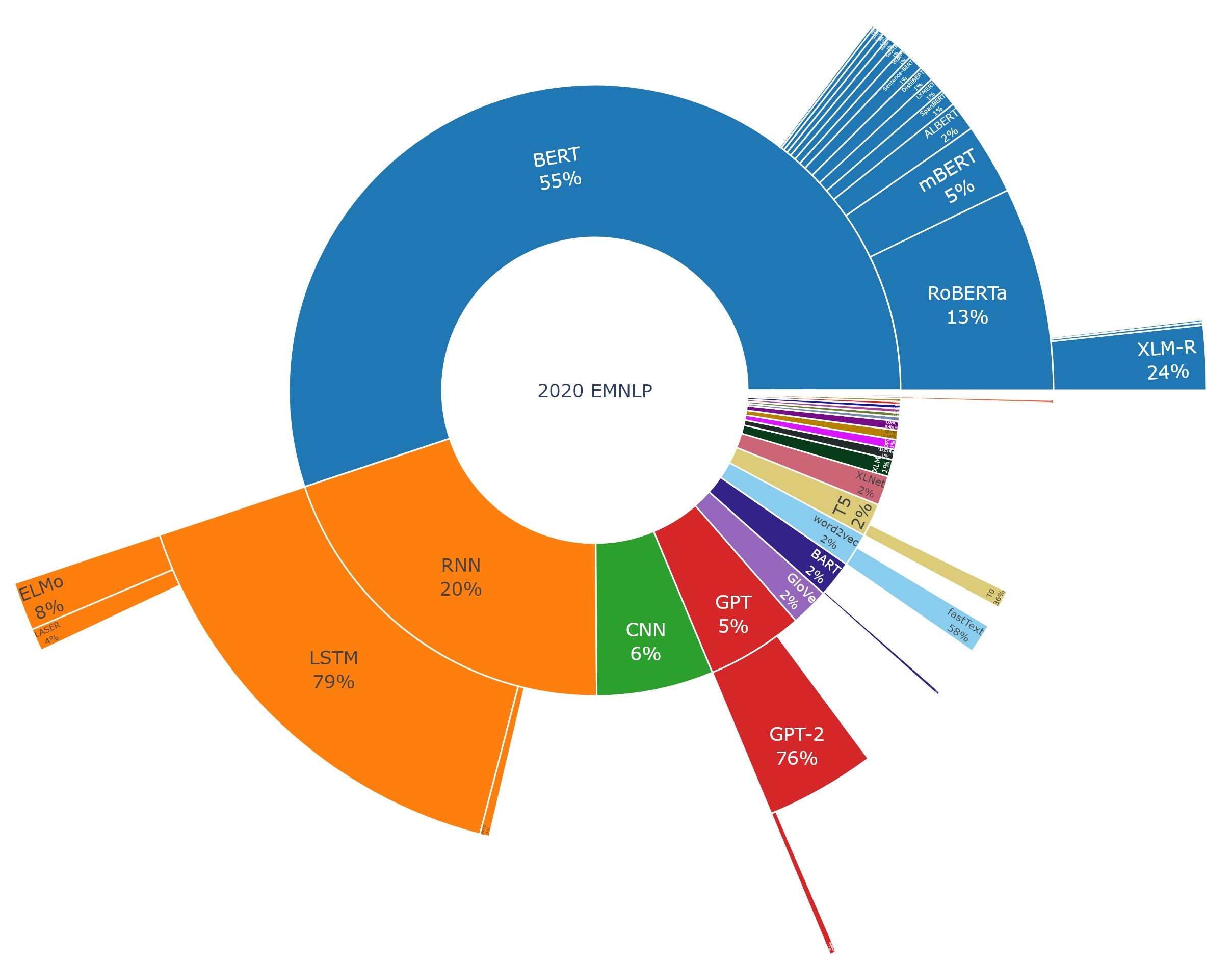}
    \includegraphics[width=\columnwidth]{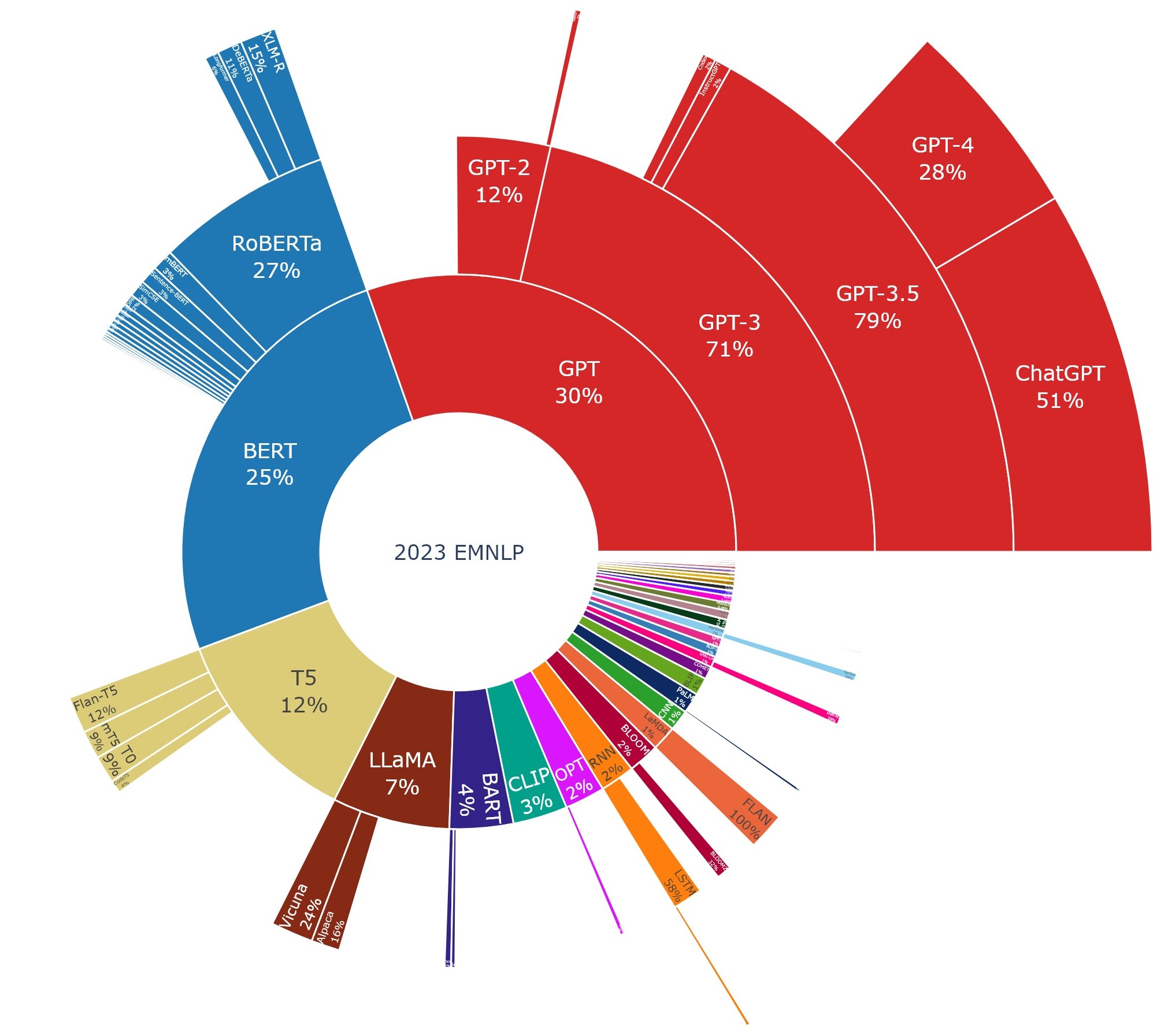}
    %\vspace*{-0.2cm}
    \caption{Compositions of model names in the main conference papers at EMNLP 2020 (left) and 2023 (right), arranged counterclockwise by their size. Branches indicate dependency, e.g., GPT-3.5 is based on and fine-tuned from GPT-3. Each group of models with the same root is represented by a unique color shared between graphs.}
    \label{fig:sunburst}
    %\vspace*{-0.4cm}
\end{figure*}

To this end, we compare how the distributions of $N^{\mathcal{L}}$ and $N$ change over time in Figure~\ref{fig:NL_vs_N}.
Each row represents a conference and the columns list all conferences which occur after it. Grid cells are pairs of conferences in comparison.
We apply a Kolmogorov-Smirnov test~\cite{massey1951kolmogorov} to each pair to determine if there is a significant difference in their distributions.
We also annotate their signed mean difference, where a positive number indicates an increased mean value from the row-conference to the succeeding column-conference. The grids are colored based on test significance level and sign of mean difference (note that all colors other than the lightest correspond to $p<0.05$).

First, we see evidence supporting our prior observations on the patterns of $N^{\mathcal{L}}$.
Earlier conferences form a cluster where no significant difference is noted;
yet, starting 2022, every conference has a significantly higher ${N^{\mathcal{L}}}$ than most or even all of its predecessors. However, the distributions of $N$ tell a distinct story. For most pairs, there is little or no evidence for a difference in distribution. There also isn't a similar line that divides the earlier and most recent conferences. For instance, the distribution of $N$ for EMNLP in 2023 does not have a significant difference with that in 2020, despite all its specialties. In fact, we even see an \textit{opposite} case: conferences in 2022 and 2023 -- the exact time of the super-linear boosts of $\bar{N^{\mathcal{L}}}$ -- contain significantly \textit{fewer} model mentions than before.
In other words, we arrive at the conclusions ``faster'': the information conveyed via specific models has not increased, but more is drawn about LMs collectively. Various factors might account for this trend: It might be that we are increasingly proficient in maintaining the Ship, or that more weight is put on the holistic concept of the Ship compared with closer examinations on the level of specific models, etc.
More effort would be needed to discern the exact cause, and a deeper understanding of the local context of LMs (e.g., similar to the work of \citealp{jurgens2018measuring}) would be especially relevant for future work.
%it could be that we the crew are increasingly proficient in replacing the planks, while it's also possible that we're fascinated more about the Ship's concept and prospect, and no longer use as much caution maintaining it in the real world.

\begin{figure}[t!]
    \centering
    \includegraphics[width=0.83\columnwidth]{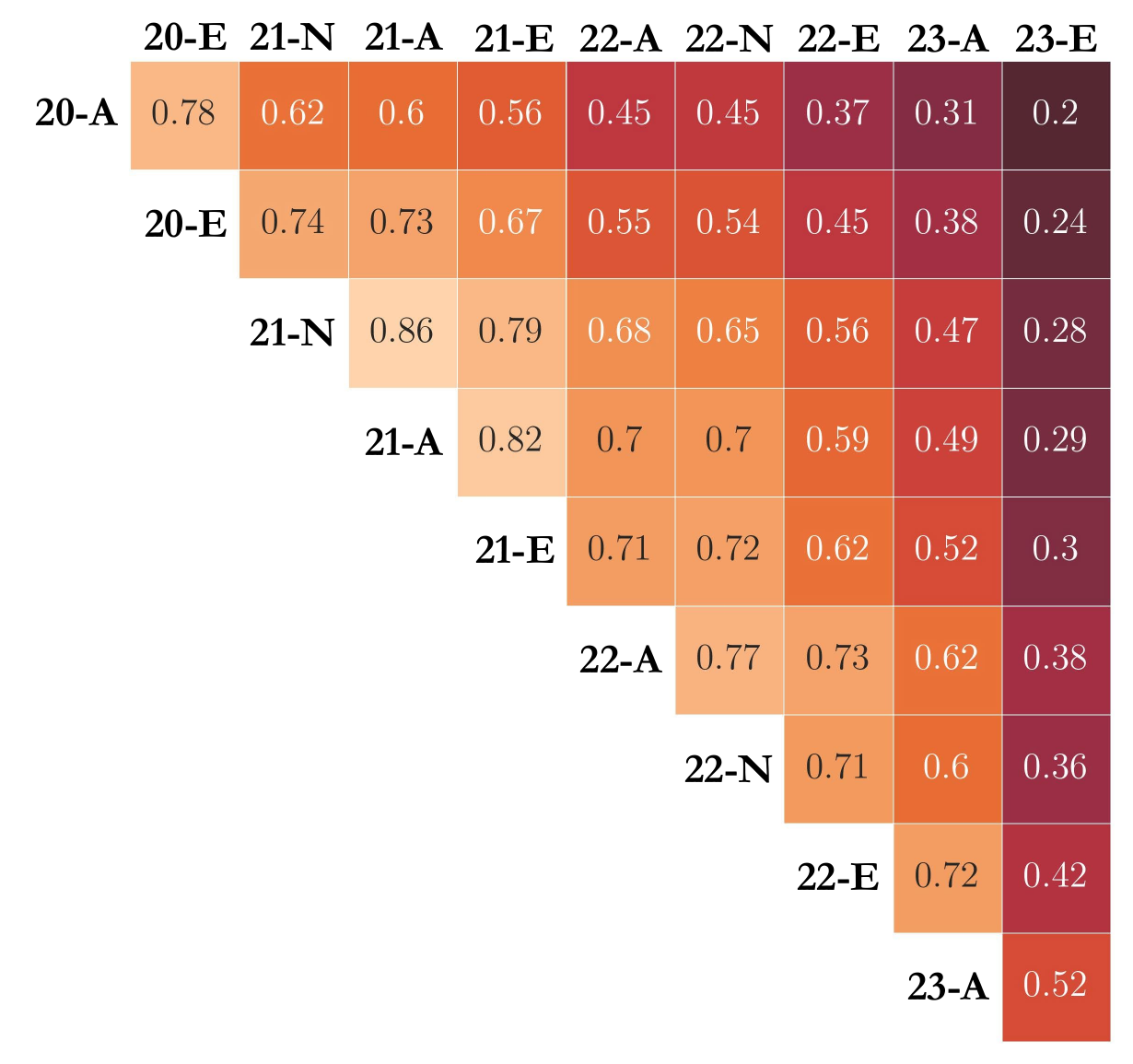}
    %\vspace*{-0.2cm}
    \caption{Jaccard similarity of model compositions between all pairs of conferences.}
    \label{fig:jaccard}
    %\vspace*{-0.4cm}
\end{figure}

\begin{figure*}[th!]
    \centering
    \includegraphics[width=0.88\columnwidth]{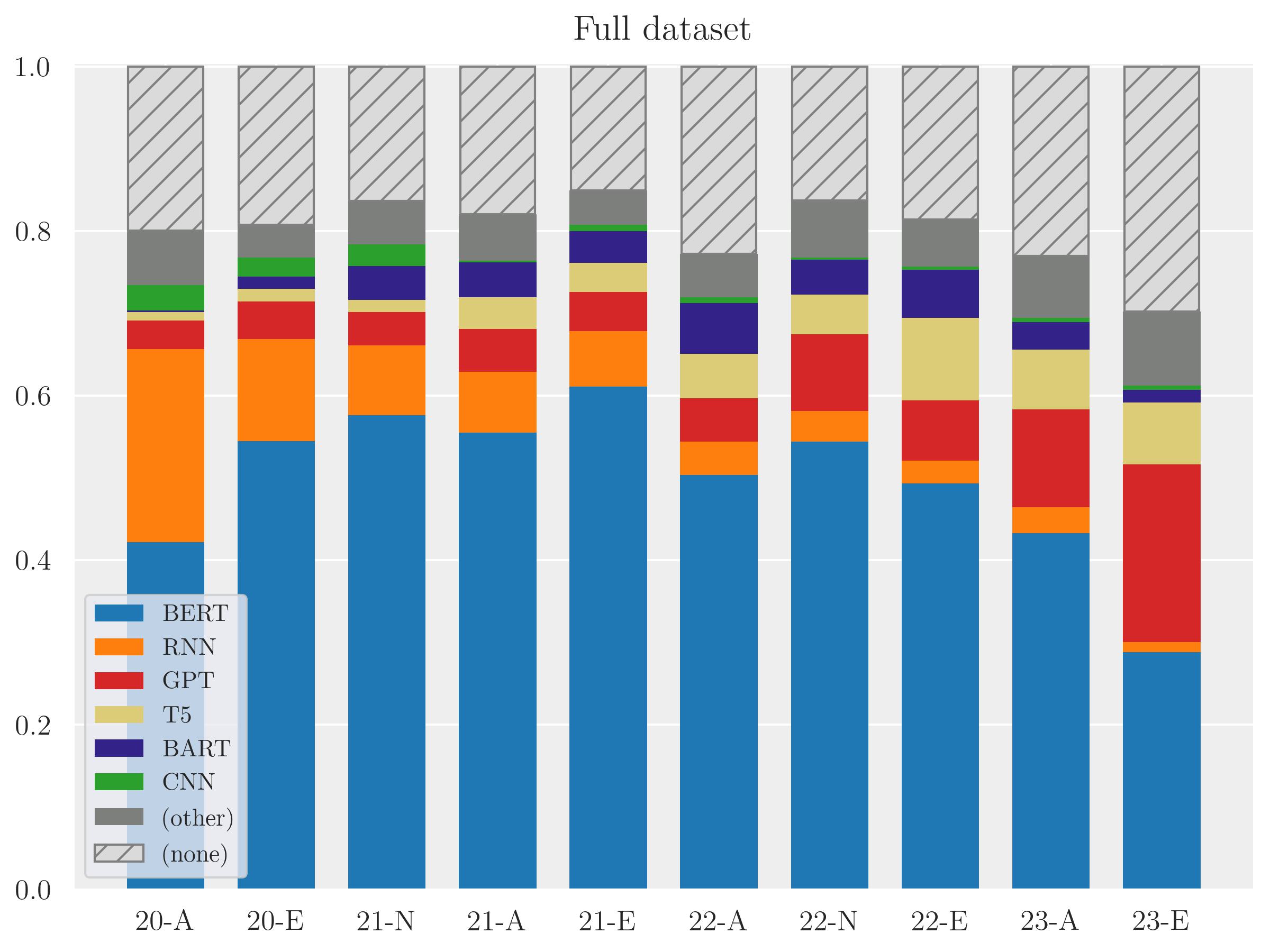}
    \includegraphics[width=0.88\columnwidth]{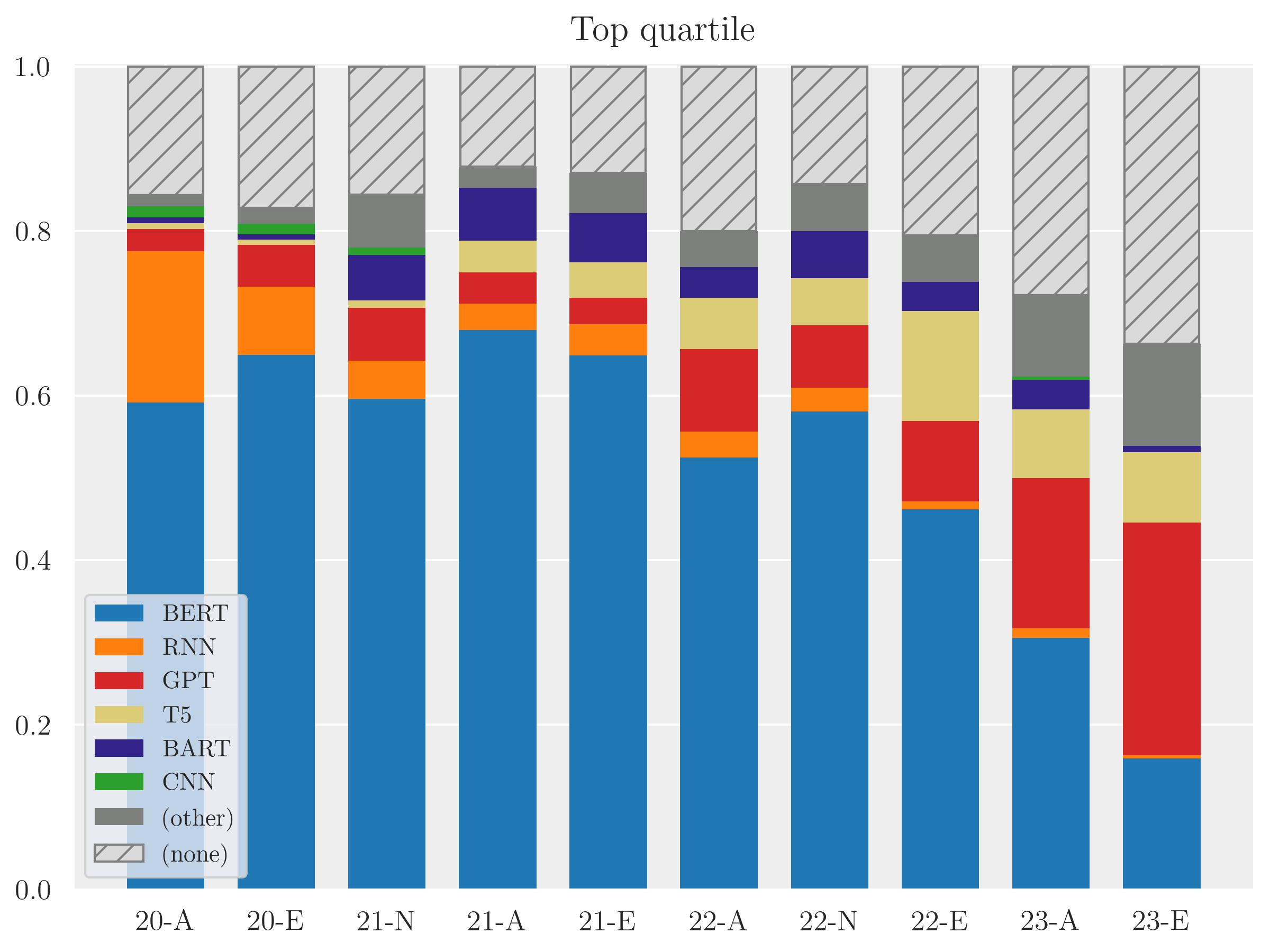}
    %\vspace*{-0.2cm}
    \caption{The existence of an \textit{absolute majority} in a paper, measured on all papers at a conference (left) and the top quarter(25\%) with highest $N^{\mathcal{L}}$ (right). The largest components are shown in the same color as Figure~\ref{fig:sunburst} and the rest are collected as ``others''. Around 80\% of papers possess an absolute majority component, and the ratio is not lowered in the most LM-centered group.}
    \label{fig:dominance}
    %\vspace*{-0.4cm}
\end{figure*}

\subsection{Oak, Pine, or Cedar Planks: Which models are we talking about?}
\label{subsec:which-models}

With the exploding usage of the LM terms comes wider variation in the use cases and context around them. To go deeper, we must consider what writers \textit{refer} to when they include LM in a paper (i.e., quantify what the Ship is like and how it's being updated at a certain point, and not just whether it sails.)
Based on all individual $N_m$ and the hierarchy of components, we obtain the exact number of the appearances of each model by matching their aliases in the text. Thus, we put together the collective compositions for each conference and visualize them as Sunburst charts, where the component sizes correspond to their share. We show a representative comparison of EMNLP 2020 and 2023 in Figure~\ref{fig:sunburst}, and display full results in Appendix~\ref{appendix:sunburst_full}.

In 2020, the BERT model alone makes up 41\% of $N$, and 55\% with its dependents (We refer to a model and all its dependents as a \textbf{component} to distinguish a group/family of models from the root model itself.) Other significant components include RNN (20\%), CNN (6\%), and GPT (5\%).
As for 2023, the GPT component (30\%) takes the lead with the advent of the notable GPT-3 models (which formed 71\% of all GPT mentions). BERT models are still the 2nd largest component despite a reduction to 25\%.
The results seems to suggest a less unipolar composition; in fact, the share of the BERT component in 2020 is comparable to the top two in 2023 combined.
We also notice the rise of more recent components, including T5 (12\%) and LLaMA (7\%), while RNN ($20\%\rightarrow2\%$) and CNN ($6\%\rightarrow1\%$) saw the most significant decreases.

How much remains as the replacement of earlier components goes on? We calculate the Jaccard similarity between compositions of conferences to quantify how much is shared across any two conferences, shown in Figure~\ref{fig:jaccard}. We observe that Jaccard similarity between conferences monotonically decreases for subsequent conferences, which matches the Ship of LMs case where its parts change over time. For two consecutive conferences, the Jaccard similarity is usually only 71\% to 86\%; the index quickly drops to 45\% to 56\% with an interval of just two years, and to 24\% to 31\% in three years. We also note that the dissimilarity is rapidly increasing, with EMNLP 2023 sharing a 52\% Jaccard similarity with ACL 2023 just half a year ago, 42\% with EMNLP in 2022, and $\le$38\% with all other predecessors. With the representing models thoroughly reshuffled in as short as $<$5 years, the ``shelf life'' of our conclusions and knowledge of LM has seen a new low, thus bringing unprecedented challenges for long-term tasks and literature studies.

\paragraph{One dominant model or many contributors?}
%\label{subsec:giants}

We have seen the presence of major component models so far,
and readers likely have their own tacit understandings of the ``giants'' in the field at the moment.
Here, we emphasize the vast implications of the dominant referents of LMs.
For example, if the supposedly abstract and inclusive concept of LM is implicitly equated with a certain model, we might be assigning the random, quirky traits of the model to the concept of LMs as a whole. This could unwittingly hinder the diversity, generalizability, and future usefulness of work despite a general veneer of neutrality among papers.

To portray how the giants shape our reported findings, we drill down to investigate their presence in \textit{individual papers}.
We examine the existence of an \textit{absolute majority} model component in each paper that appears more than all other components combined, i.e., occupying more than $N/2$.
One scenario, then, would be that a single or small set of giants actually underpin the notion of LMs in papers. On the other hand, if LM is truly a general term of art, then we might also see some but not most papers dominated by a model.

Figure~\ref{fig:dominance} displays the proportion of publications with absolute majority components for the full data (left) and the top 25\% with the highest $N^\mathcal{L}$ (right). The most notable components are marked with the same color as in Fig.~\ref{fig:sunburst}. Other models are collected as the grey bar (``others''), and the proportion with no absolute majority is denoted with a striped pattern. Around 80\% of the papers contain an absolute majority model. Specifically, we get a glance at the astounding traction of BERT before more recent paradigm shifts: it dominated up to 61.1\% of all papers and 67.9\% of the most LM-centered ones.

Interestingly, more focus on the collective LM terms did not entail a more balanced composition. In fact, they are often more biased: The percentage of papers with an absolute majority is \textit{higher} in the most LM-centered quartile than overall for all 7 conferences before EMNLP 2022. There has also been a fresh wind, however. In the most recent 3 conferences, we see \textit{fewer} dominant components when a paper focuses heavily on the generalized LM terms. More importantly, the chance of an absolute majority in both groups has been on a continuing decline and both reached an unprecedentedly lower level: 70.2\% for all papers and 66.3\% for the top quarter. We call to keep monitoring the heterogeneity (or lack thereof) in LM papers given the (still) high presence of major components and a visibly surging presence of the GPT component at the most recent EMNLP 2023.

\subsection{Lembos or Trireme: Factoring in context}
\label{subsec:diff_among_groups}
The extent to which a paper focuses on LMs implies different use scenarios. For instance, a work may utilize and mention them for data processing but doesn't concern the science of LMs per se. This usually implies lower $N^\mathcal{L}$ in contrast to, say, another work that studies an emergent property of LLMs. As we explore how LMs are embodied in papers, we need to consider how the composition of papers factors into the usage of ``LMs''.

\begin{figure}[t!]
    \centering
    \includegraphics[width=\columnwidth]{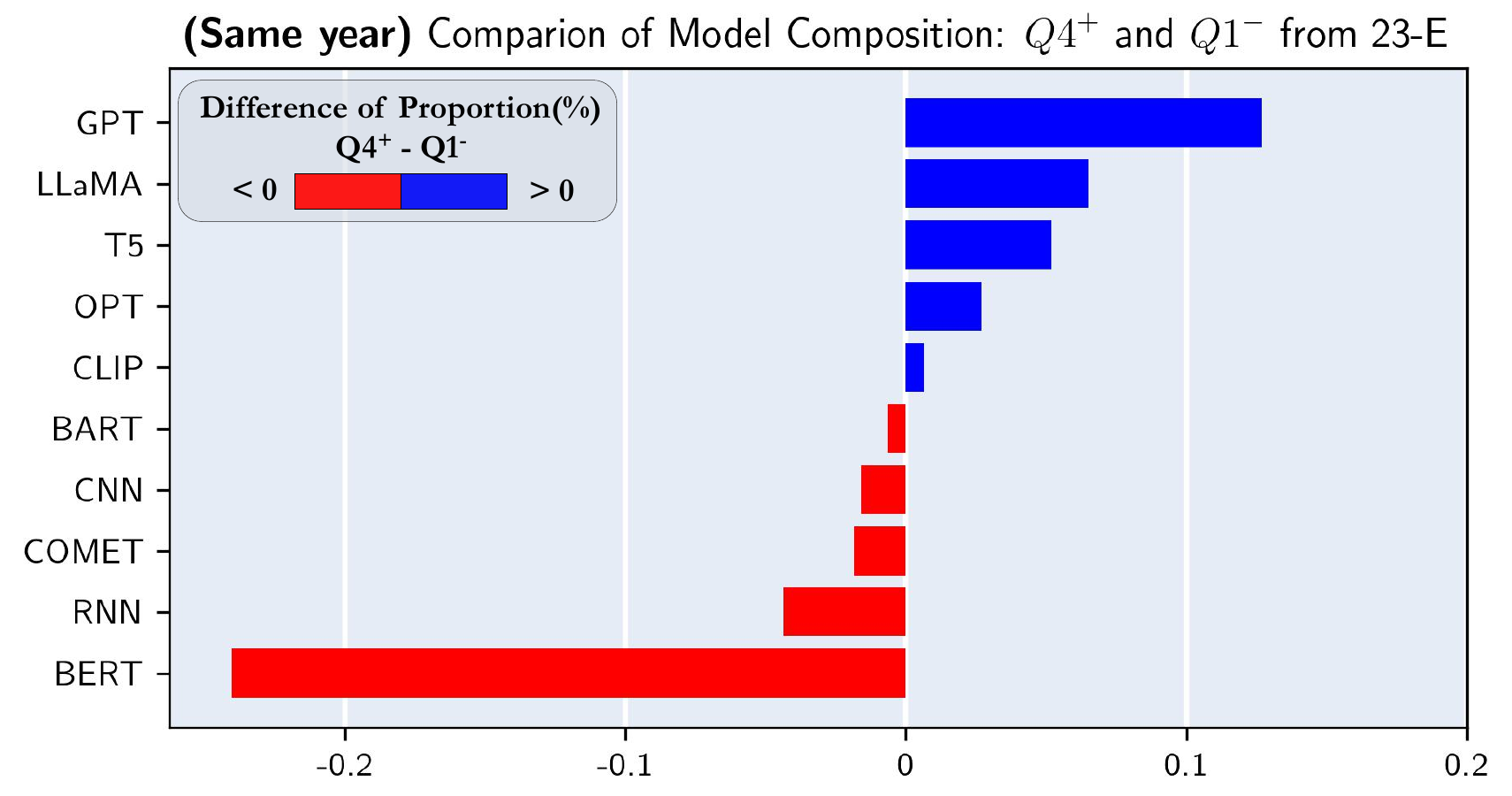}
    %\vspace*{-0.2cm}
    \caption{Differences in the model composition of the most LM-focused quarter (${Q4}^{+}$) and the least one (${Q1}^{-}$). Blue/Red denotes a higher/lower share in ${Q4}^{+}$.}
    \label{fig:same_year_quartile_comparison}
    %\vspace*{-0.3cm}
\end{figure}

We rank the LM-related papers at a conference by $N^\mathcal{L}$ and compare between the ones with the most and least focus on the LM terms. Specifically, we extract the top and bottom quarters, denoted as ${Q4}^{+}$ and ${Q1}^{-}$ respectively. Figure~\ref{fig:same_year_quartile_comparison} compares compositions of $N$ from ${Q4}^{+}$ and ${Q1}^{-}$ of EMNLP 2023 and shows the differences in the 10 largest components. In ${Q4}^{+}$, the GPT component covers an additional 12.7\% of $N$ compared with ${Q1}^{-}$, followed by LLaMA (6.5\%) and T5 (5.2\%). Meanwhile, the BERT component takes as much as an extra 24.0\% in ${Q1}^{-}$, where the roles of the LMs terms are most peripheral.

\begin{figure}[t!]
    \centering
    \includegraphics[width=\columnwidth]{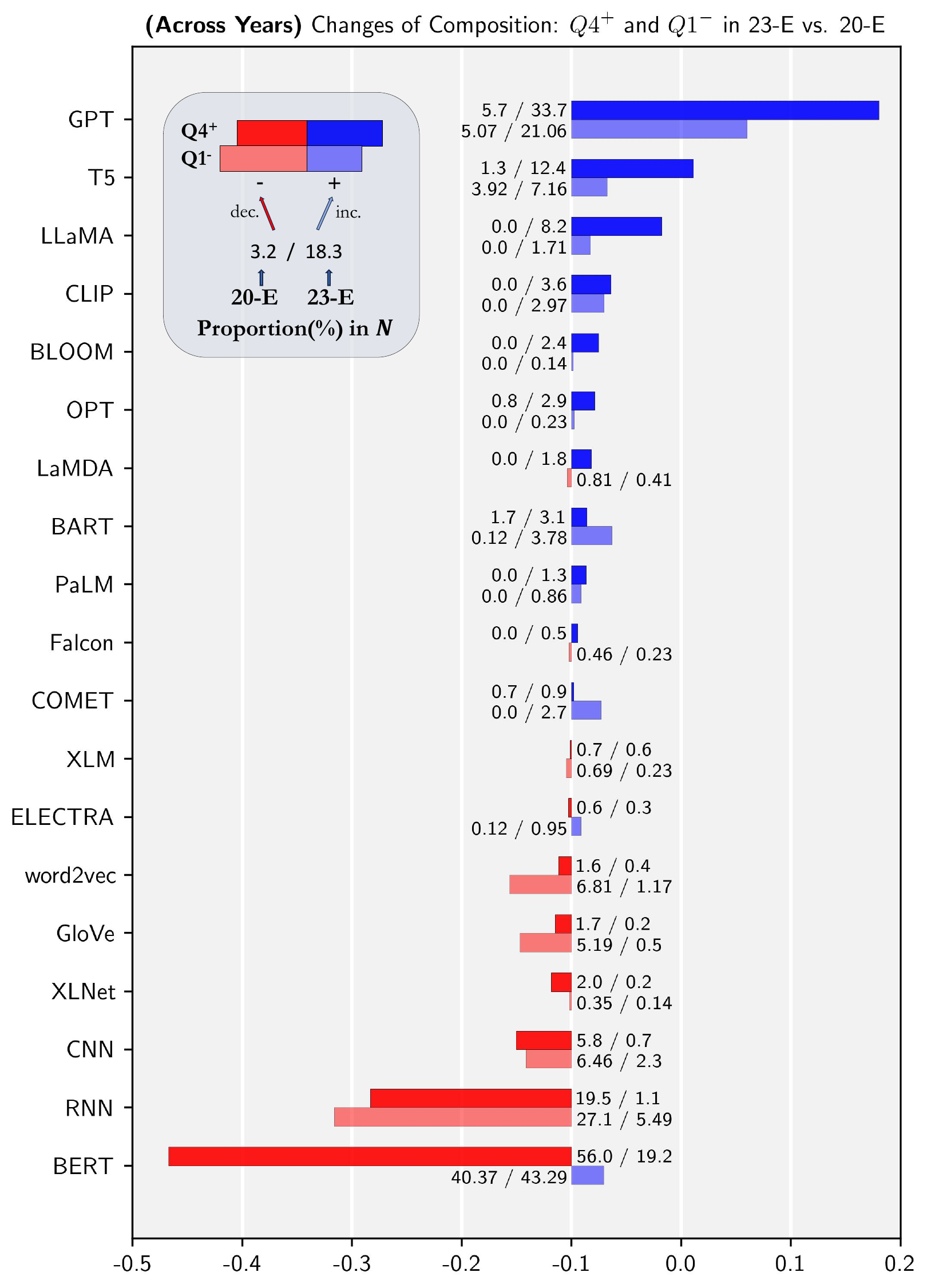}
    %\vspace*{-0.2cm}
    \caption{Differences in the model composition of EMNLP 2020 and 2023, measured respectively on the most LM-focused quarter (${Q4}^{+}$; darker colors) and the least one (${Q1}^{-}$; lighter colors), and ranked by the share of ${Q4}^{+}$ in 23-E. Blue/Red denotes an increase/decrease of share. We annotate beside each bar the exact numbers in 2020 and 2023, separated by `/'.}
    \label{fig:over_years_quartile_comparison}
    %\vspace*{-0.4cm}
\end{figure}

We further measure how \textit{the same group changes over time}, comparing model compositions at an early conference (EMNLP 2020) versus a most recent one (EMNLP 2023). We depict the change of 15 largest components in Figure~\ref{fig:over_years_quartile_comparison}. The findings are consistent: the most LM-centered group adapts thoroughly to the latest models, while ${Q1}^{-}$ sees a much more modest change, or even increased interest for models pre-2020 like BART~\cite{lewis-etal-2020-bart} and COMET~\cite{bosselut-etal-2019-comet}.
Most notable is the contrast for BERT: For ${Q4}^{+}$, the once-dominant component loses as much as 36.8\% of $N$, going from 56.0\% to 19.2\%. However, in ${Q1}^{-}$, the BERT component remains unaffected: it still firmly takes up more than 40\% of $N$, and even saw a $\sim$3\% \textit{increase}, despite all the new model components and families.
This demonstrates that the trend in less LM-centered groups is not merely a moderated or delayed version of that in the top ones, but indeed represents a distinct interpretation and resolution of LMs.

The findings converge to an interesting division: the newest components or those with the latest major models (GPT-3, ChatGPT, LlaMA, and Flan-T5, inter alia; all of which are post-2021) instantly enter the most LM-centered discourse, while the least LM-centered ones persist to favor certain earlier models for a longer period.
Thus, we should be aware of the chasm between the co-occurring yet distinct contexts that eventually map to different constructions of the same Ship. For instance, a novel property found with 2023 models as default could be problematic if communicated to a BERT-centered subfield without regard, and further used to justify the use of LMs in a new stake despite wildly different ship drafts and capabilities. In return, the context difference may further hinder communications between groups representing the varied use and interpretations of LMs.

\section{Concluding Remarks}
\label{sec:discussion}

In this work, we go over the past and present of the enduring term of Language Model(s), based on an original dataset from the latest major conferences. We sort out the subtle, continuous shifts in the practical meaning of LMs, and witness how the retrofits eventually accumulate to a brand new \textit{Ship of Language Models} and the nuances in the actual referents. We quantify and visualize the drastic change in the planks and timber, emphasizing the shortened period of reconstruction and the presence of dominant components. Finally, we highlight the snowballing context difference between the LM-centered research and the more peripheral applications and the consequences of that.

Approaching the most recent epochal crossroads where revolutionary shifts can happen within months or weeks, we have in fact seen more interesting signs: the unprecedented dissimilarity of compositions, or on the other hand, a more diverse, multipolar representation in model choices, to name a few. Perhaps what is crafted from our hands has even brought up throught the Age of Sail, where a now steam-engined Argo is ready to set sail towards the exciting yet mysterious fanta-seas.

\paragraph{Future Directions}
Our findings depict the most dramatic reshuffles of \textit{models}, while the nuances of LMs can be found similarly at the underlying \textit{architecture} level, or different \textit{variations} of models.
Besides, causal links between our work and the identified key factors in prior works may help depict the field's trajectories, and qualitative studies such as interviews with paper authors can also help understand the different patterns and beliefs.
Beyond the LM-specific discussions, we also highlight the broad existence of the Ship of Theseus scenario in various other terms, topics, and fields. 
For instance, it would be intriguing if similar analyses can be applied to understand how the interpretations of \textit{fairness} or \textit{safety} in the AI context were enriched over time.

\section*{Limitations}
While our methodologies can be naturally extended, we would like to note that the current analyses and implementations have limitations.
Although we reviewed various findings in recent years with the most dramatic advances of LMs, a holistic overview of a research topic, let alone an entire field, is not covered by the scope of a few years. It would be meaningful if these most rapid changes could be connected with the decades of conceptualization and exploration preceding the engineering breakthroughs.
Similarly, using main conference papers at the major international venues as a proxy of the NLP community has its limitations. We encourage future works to take broader consideration of the essential contributions that are less represented by the relatively convergent selections of such venues, e.g., regional conferences on dialect or indigenous language.

More broadly, text-based methods alone are not sufficient to cast the intricate dynamics of science. Scientific communities are not mere carriers of printed works, and the influence of a language model or a paper is far beyond academic language use. Various other important factors and impacts should be considered for a comprehensive description of scientific progress and any specific scientific products: the status quo of subfields, monetary and environmental costs of implementation, societal impacts and the public's perception, etc.

\section*{Ethical Considerations}
Our data is collected from the ACL Anthology on the terms of Creative Commons 4.0 BY (Attribution) license, which allows unlimited reproduction, distribution, and hosting of materials for non-commercial purposes\footnote{See \href{https://creativecommons.org/licenses/by/4.0/}{CC BY 4.0} and \href{https://aclanthology.org/faq/}{the ACL Anthology's copyright statement} for more information.}. The authors report no other potential ethical considerations.

\bibliography{custom}

\appendix

\section{The dataset construction pipeline}

We provide more details about the semi-automated dataset creation process, including the full list of model names, and the prompts and parameters for the automated part of model name identification.

\subsection{A note on the scope of ``LMs''}
\label{appendix:LM_definition}

In the rigid sense, LMs shall correspond directly to the task of next-word prediction by enlisting the probabilities of each possible next-word (adapted from the prose of \citet{jurafsky2023LMdef}).
However, practical referents and contexts of “language modeling” in the real world have been broad and complex. In fact, in Jurafsky and Martin’s work cited above, the definition of LMs (given as a direct statement) has already been broader than previous senses in the prose:

\begin{quote}
    Models that assign probabilities to upcoming words, or \textbf{sequences of words in general}, are called language models or LMs.
\end{quote}

To accommodate the underspecification and the even more vague real-world usages, we thus consider the ``broadest'' definitions that are historically involved. This would include text embedding schemes that have essentially been used in the place of “LMs”, key parts of an LM-like system where the system itself is not otherwise named, or artifacts created in the “LM” way but are not exactly a “model”. (Note that this work concerns ``\textit{what} we talk about when we talk about LMs'', not just ``\textit{which LM(s)} we talk about when we talk about LMs''.)

This is especially worth noting since the terminology of ``LM'' is not always the most prevalent. As an important instance, “word embeddings/vectors” had been a central topic; it is both true that (1) the terms are not equivalent to ``LMs'' strictly speaking, and that (2) it was a very common (if not the default) practice to build (e.g. add a linear layer) on top of the embeddings to obtain probabilities/likelihood of the text in question. This is essentially close to “LM” as defined, though we rarely referred to the entire system this way but would have often used wordings like ``word vectors + [classification/prediction/…]'' back then. To this end, we do consider these ``broader'' senses because they have long been used for the same sense and in the same context. In fact, our motivations align with such situations: the communities’ description and perception of a concept might have undergone many changes and variations, but we are yet to know what exactly has changed, to what extent, and how varied it is.

We exemplify our considerations below with some keywords in the list:

\begin{itemize}
    \item Word2vec is the algorithm where the training goal is exactly to assign a max likelihood/probability of the current text (and not to forget N-gram LMs is indeed a type of LM too -- there goes the Ship of LMs!);
    \item The reasons for other word embedding schemes like GloVE has been addressed by the discussions so far;
    \item RNN, LSTM, etc. are the neural networks that generate continuous text representations, and they have been prominently applied for the exact task of probabilistic next-word/sentence generation.
    \item There are also individual considerations. For instance, CNN, CLIP, and wav2vec have been under discussion regarding the multimodality issues, and we eventually decided to keep (potentially) multimodal keywords.
\end{itemize}

%\subsection{Classifying potential model names}
%\label{appendix:fig-classify_workflow}
%See Figure~\ref{fig:workflow}. 

\subsection{Full list of models}
\label{appendix:full_model_entry}
ChatGPT, GPT-3, GPT-4, BERT, T5, GPT-3.5, GPT-2, LLaMA, RoBERTa, PaLM, CLIP, BART, XLM-R, Alpaca, BLOOM, mT5, InstructGPT, mBERT, GPT-J, Flan-T5, OPT, Codex, COMET, ELECTRA, Longformer, mBART, SimCSE, BLOOMZ, BigBird, BLIP, DeBERTa, CodeT5, Switch Transformer, Vicuna, T0, PEGASUS, LSTM, ALBERT, DPR, Macaw, LXMERT, SpanBERT, TinyBERT, ViLBERT, XLM, Linformer, kNN-LM, kNN-MT, REALM, RETRO, GraphCodeBERT, Sentence-BERT, RNN, HyperCLOVA, CodeGen, Dolly, Pythia, LaMDA, FLAN, BLIP-2, XLNet, GPT, ELMo, BioBERT, DialoGPT, RemBERT, PaLM 2, DistilBERT, SciBERT, ClinicalBERT, M2M100, GloVe, LASER, word2vec, fastText, LaBSE, CNN, wav2vec, UNITER, WizardLM, MASS, MT-DNN, BlenderBot, OFA, CMLM, HRED, ERNIE, ConveRT, Diffusion-LM, MiniLM, Falcon, Galactica, PPLM, RuleTakers, Claude, Tk-Instruct, LayoutLM, PanGu-$\alpha$, GROVER, CTRL, EntityNLM, ST-DNN, SpERT

\subsection{Settings of the automated model name detection module}
\label{appendix:LLM_setup}
The LLM-based name detection is performed with OpenAI's \texttt{gpt-4-0125-preview} (also known as \texttt{gpt-4-turbo-preview})\footnote{\url{https://platform.openai.com/docs/models\#gpt-4-turbo-and-gpt-4}}, accessed in February to June 2024. The temperature is set to 0, and we follow the default settings of other parameters, e.g., top\_p = 1.00.

\subsection{Prompts for automated model name extraction}
\label{appendix:prompt}
We use GPT-4-turbo as our base LLM to identify potential names from paper abstracts and incorporate in-context examples~\cite{dong2022survey}. A request consists of two parts of inputs: a static \textbf{System} instruction, and individual \textbf{User Inputs} for each request. An example use case is shown below:

\paragraph{System} 

\

You are an assistant with excellent expertise in searching through academic text. You will be given the Abstract of an academic paper in the field of Natural Language Processing. Your task is to retrieve whether the authors mention that they mentioned some *specific* language model in their writing. And if so, you need to accurately find the names of all such models.

Important note 1: "LLM" and "PLM" are not model names, they refer to the generic terms of "Large Language Model" and "Pretrained Language Model".

Important note 2: Do not include any models that are proposed by the authors themselves. For instance, if a paper says "we propose a new model, GPT-OURNEW, which performs better than GPT-3", your answer should only include "GPT-3" and not "GPT-OURNEW".

Return all the specific model names (don't miss out any), separated by a comma. If you believe you didn't see any model name, simply return "None". Only respond with the comma-separated model names. Do not include any other text in your response!!!

\,

Some examples:

Input: This paper explores the potential of leveraging Large Language Models (LLMs) for data augmentation in multilingual commonsense reasoning datasets where the available training data is extremely limited. To achieve this, we utilise several LLMs, namely Dolly-v2, StableVicuna, ChatGPT, and GPT-4, to augment three datasets: XCOPA, XWinograd, and XStoryCloze. Subsequently, we evaluate the effectiveness of fine-tuning smaller multilingual models, mBERT and XLMR, using the synthesised data. We compare the performance of training with data generated in English and target languages, as well as translated English-generated data, revealing the overall advantages of incorporating data generated by LLMs, e.g. a notable 13.4 accuracy score improvement for the best case. Furthermore, we conduct a human evaluation by asking native speakers to assess the naturalness and logical coherence of the generated examples across different languages. The results of the evaluation indicate that LLMs such as ChatGPT and GPT-4 excel at producing natural and coherent text in most languages, however, they struggle to generate meaningful text in certain languages like Tamil. We also observe that ChatGPT falls short in generating plausible alternatives compared to the original dataset, whereas examples from GPT-4 exhibit competitive logical consistency.

Output: Dolly-v2,StableVicuna,ChatGPT,GPT-4,mBERT,XLMR

\,

Input: Large Language Models (LLMs) have showcased impressive performance. However, due to their inability to capture relationships among samples, these frozen LLMs inevitably keep repeating similar mistakes. In this work, we propose our Tuning-free Rule Accumulation (TRAN) framework, which guides LLMs in improving their performance by learning from previous mistakes. Considering data arrives sequentially, LLMs gradually accumulate rules from incorrect cases, forming a rule collection. These rules are then utilized by the LLMs to avoid making similar mistakes when processing subsequent inputs. Moreover, the rules remain independent of the primary prompts, seamlessly complementing prompt design strategies. Experimentally, we show that TRAN improves over recent baselines by a large margin.

Output: None

\,

Input: Dialogue State Tracking (DST) is of paramount importance in ensuring accurate tracking of user goals and system actions within task-oriented dialogue systems. The emergence of large language models (LLMs) such as GPT3 and ChatGPT has sparked considerable interest in assessing their efficacy across diverse applications. In this study, we conduct an initial examination of ChatGPT’s capabilities in DST. Our evaluation uncovers the exceptional performance of ChatGPT in this task, offering valuable insights to researchers regarding its capabilities and providing useful directions for designing and enhancing dialogue systems. Despite its impressive performance, ChatGPT has significant limitations including its closed-source nature, request restrictions, raising data privacy concerns, and lacking local deployment capabilities. To address these concerns, we present LDST, an LLM-driven DST framework based on smaller, open-source foundation models. By utilizing a novel domain-slot instruction tuning method, LDST achieves performance on par with ChatGPT. Comprehensive evaluations across three distinct experimental settings, we find that LDST exhibits remarkable performance improvements in both zero-shot and few-shot setting compared to previous SOTA methods. The source code is provided for reproducibility.

Output: ChatGPT

\paragraph{User Input}

\

Input: We propose LLM-FP4 for quantizing both weights and activations in large language models (LLMs) down to 4-bit floating-point values, in a post-training manner. Existing post-training quantization (PTQ) solutions are primarily integer-based and struggle with bit widths below 8 bits. Compared to integer quantization, floating-point (FP) quantization is more flexible and can better handle long-tail or bell-shaped distributions, and it has emerged as a default choice in many hardware platforms. One characteristic of FP quantization is that its performance largely depends on the choice of exponent bits and clipping range. In this regard, we construct a strong FP-PTQ baseline by searching for the optimal quantization parameters. Furthermore, we observe a high inter-channel variance and low intra-channel variance pattern in activation distributions, which adds activation quantization difficulty. We recognize this pattern to be consistent across a spectrum of transformer models designed for diverse tasks such as LLMs, BERT, and Vision Transformer models. To tackle this, we propose per-channel activation quantization and show that these additional scaling factors can be reparameterized as exponential biases of weights, incurring a negligible cost. Our method, for the first time, can quantize both weights and activations in the LLaMA-13B to only 4-bit and achieves an average score of 63.1 on the common sense zero-shot reasoning tasks, which is only 5.8 lower than the full-precision model, significantly outperforming the previous state-of-the-art by 12.7 points. Code is available at: https://github.com/nbasyl/LLM-FP4.

Output: 

\paragraph{Agent Response}

\

BERT,Vision Transformer,LLaMA-13B

\section{Sunburst diagrams}
\label{appendix:sunburst_full}

The following pages show the Sunburst diagrams for all 10 conferences in chronological order.

\begin{figure*}[th!]
    \centering
    \includegraphics[width=0.88\textwidth]{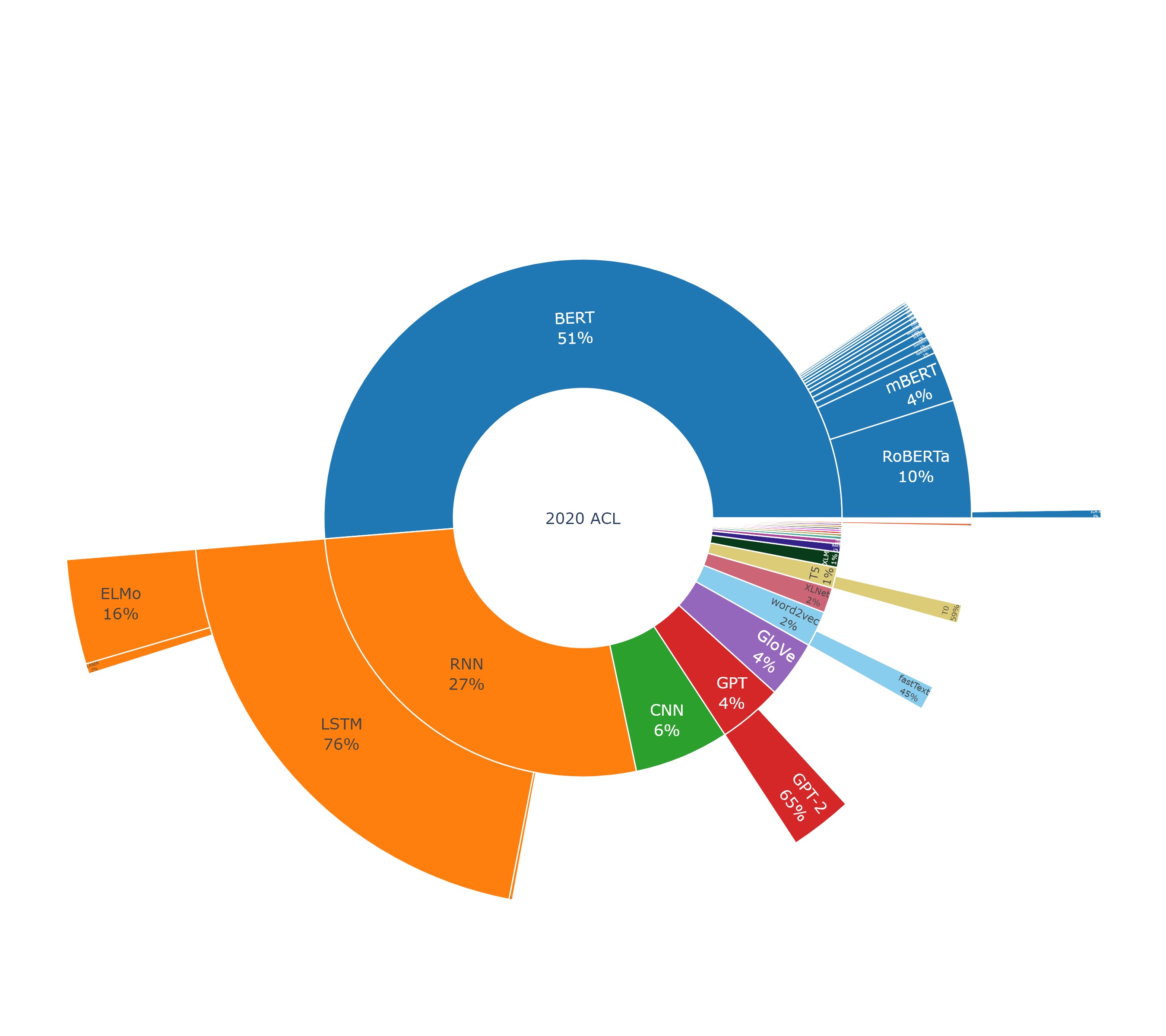}
    \vspace*{-1cm}
    \includegraphics[width=0.88\textwidth]{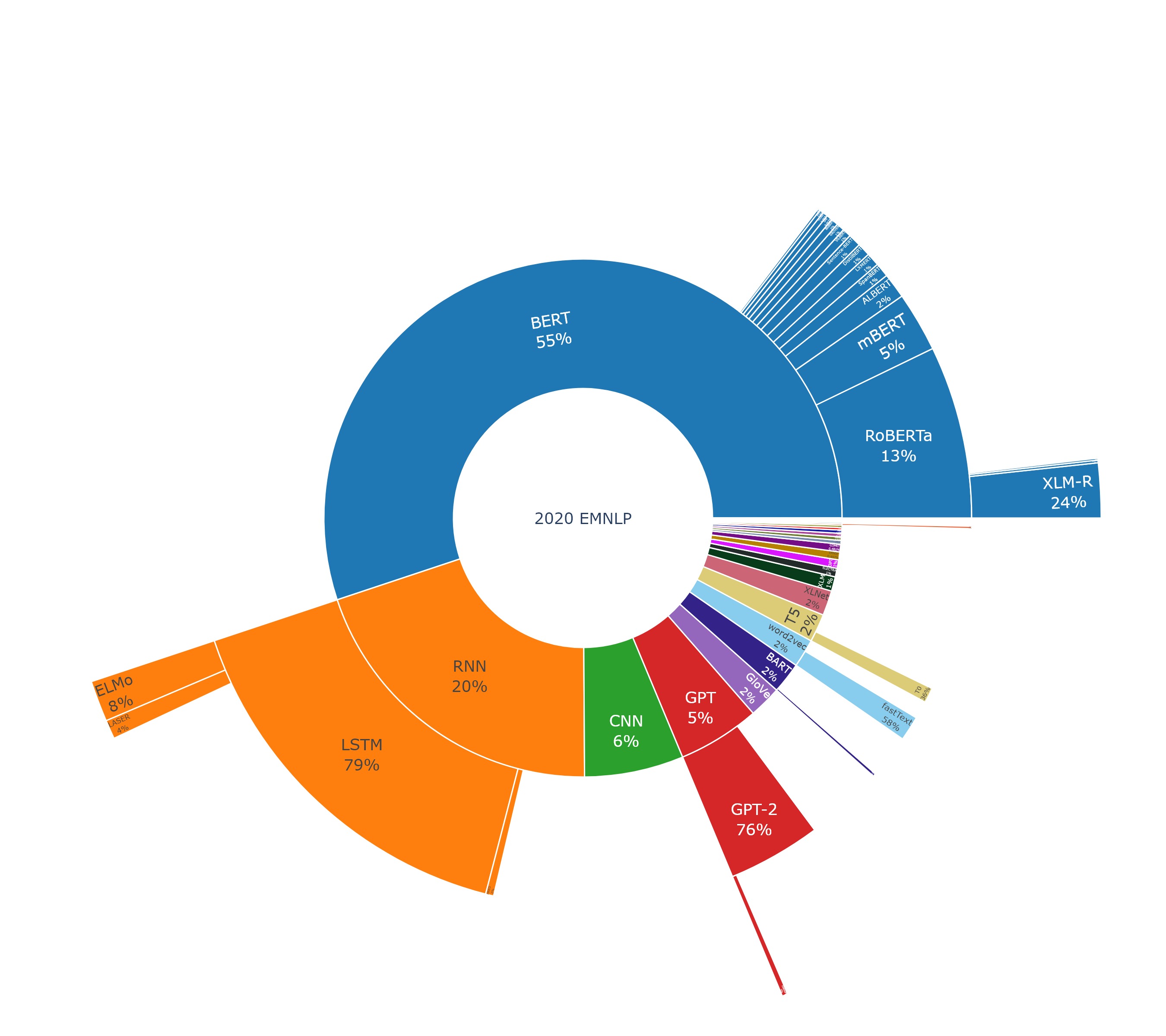}
\end{figure*}

\begin{figure*}[th!]
    \centering
    \includegraphics[width=0.88\textwidth]{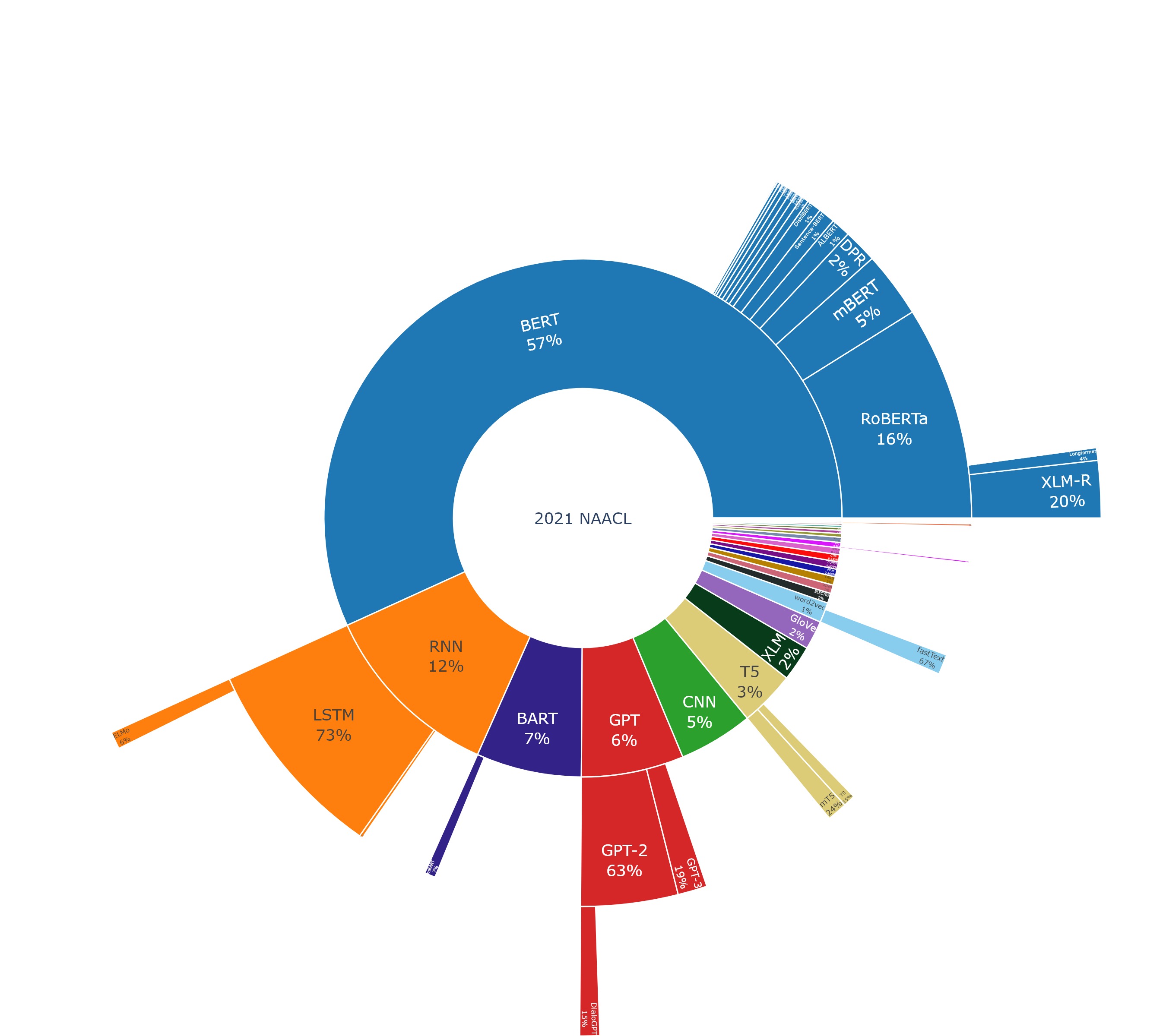}
    \vspace*{-1cm}
    \includegraphics[width=0.88\textwidth]{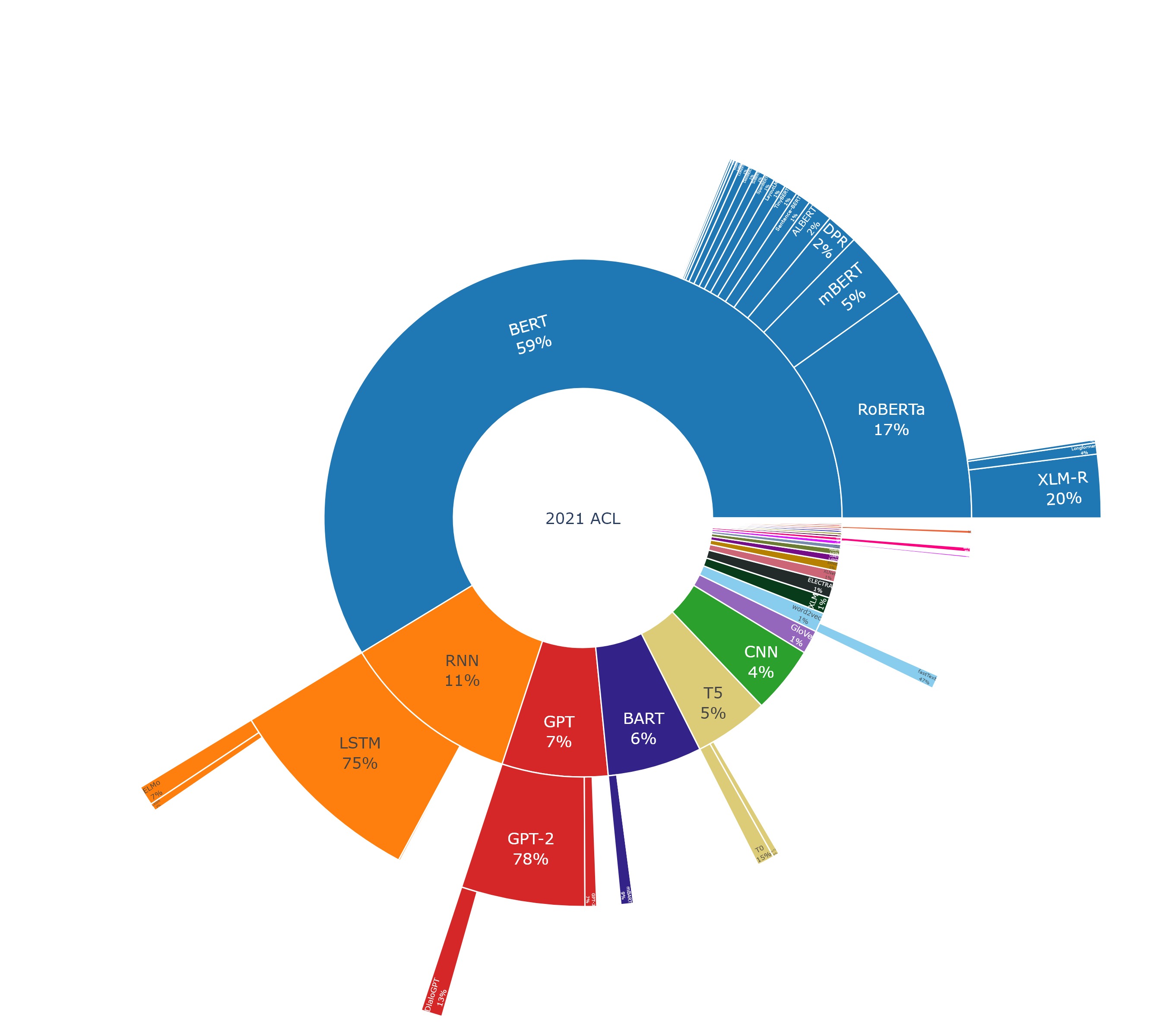}
\end{figure*}

\begin{figure*}[th!]
    \centering
    \includegraphics[width=0.88\textwidth]{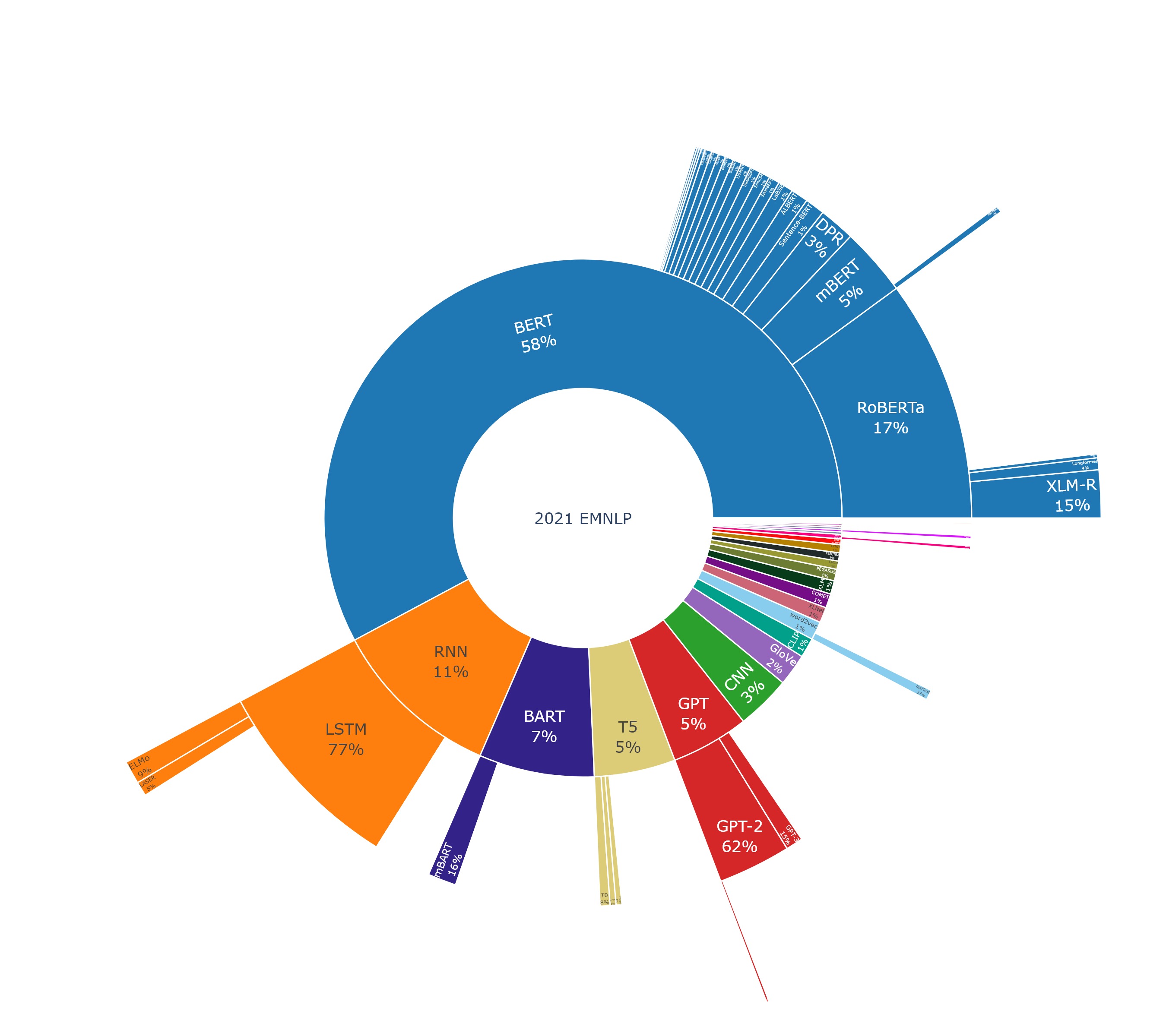}
    \vspace*{-1cm}
    \includegraphics[width=0.88\textwidth]{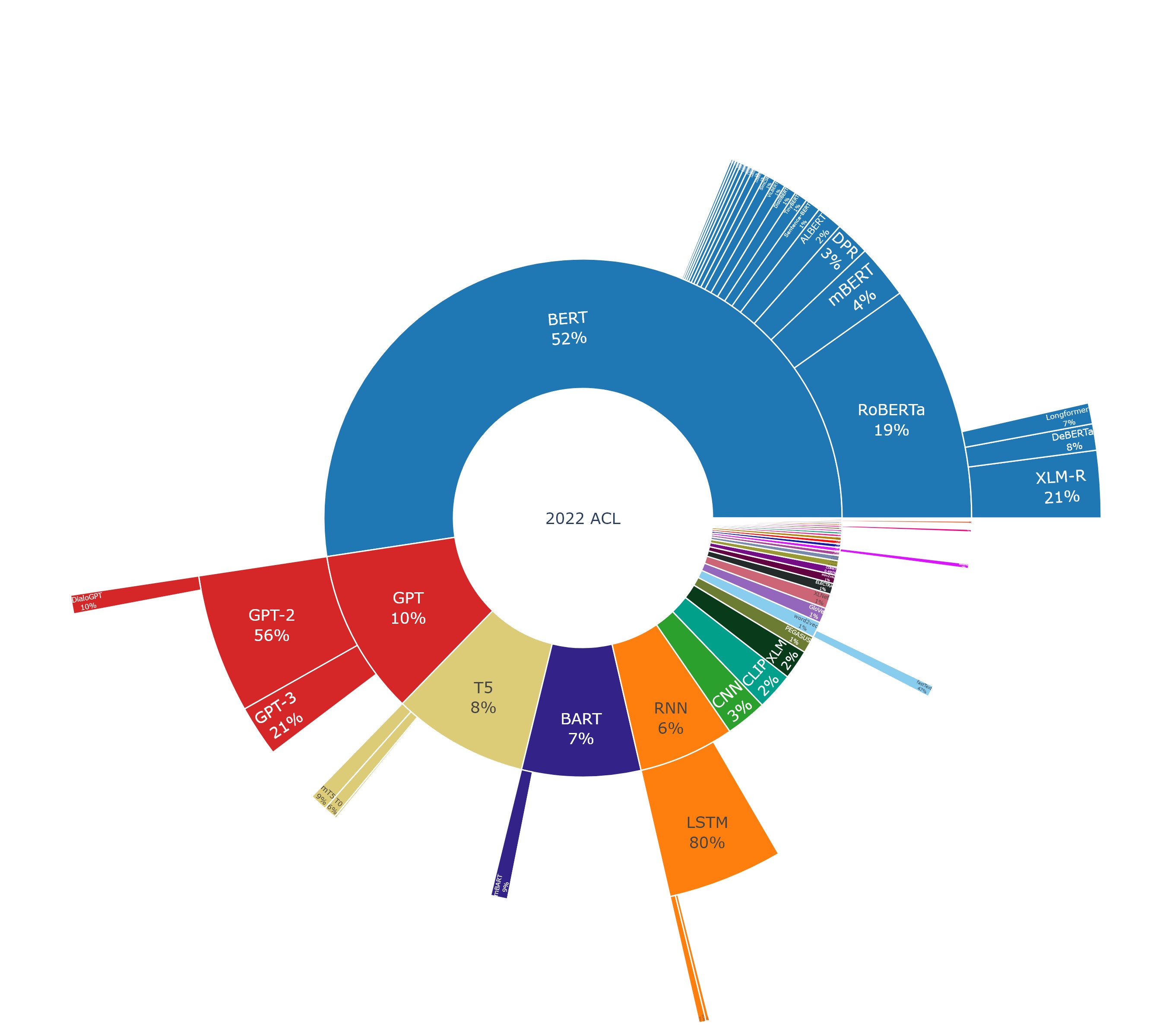}
\end{figure*}

\begin{figure*}[th!]
    \centering
    \includegraphics[width=0.88\textwidth]{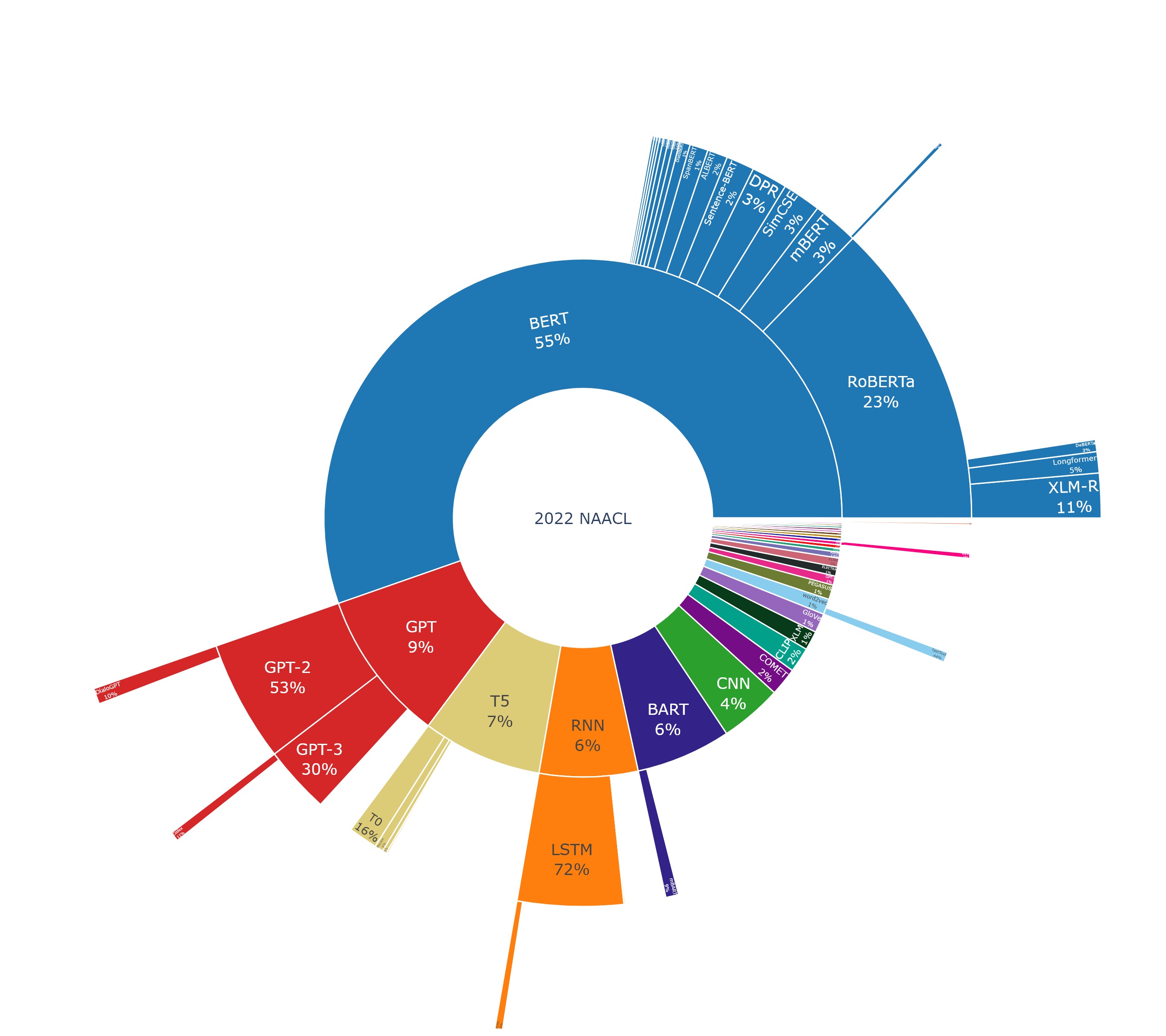}
    \vspace*{-1cm}
    \includegraphics[width=0.88\textwidth]{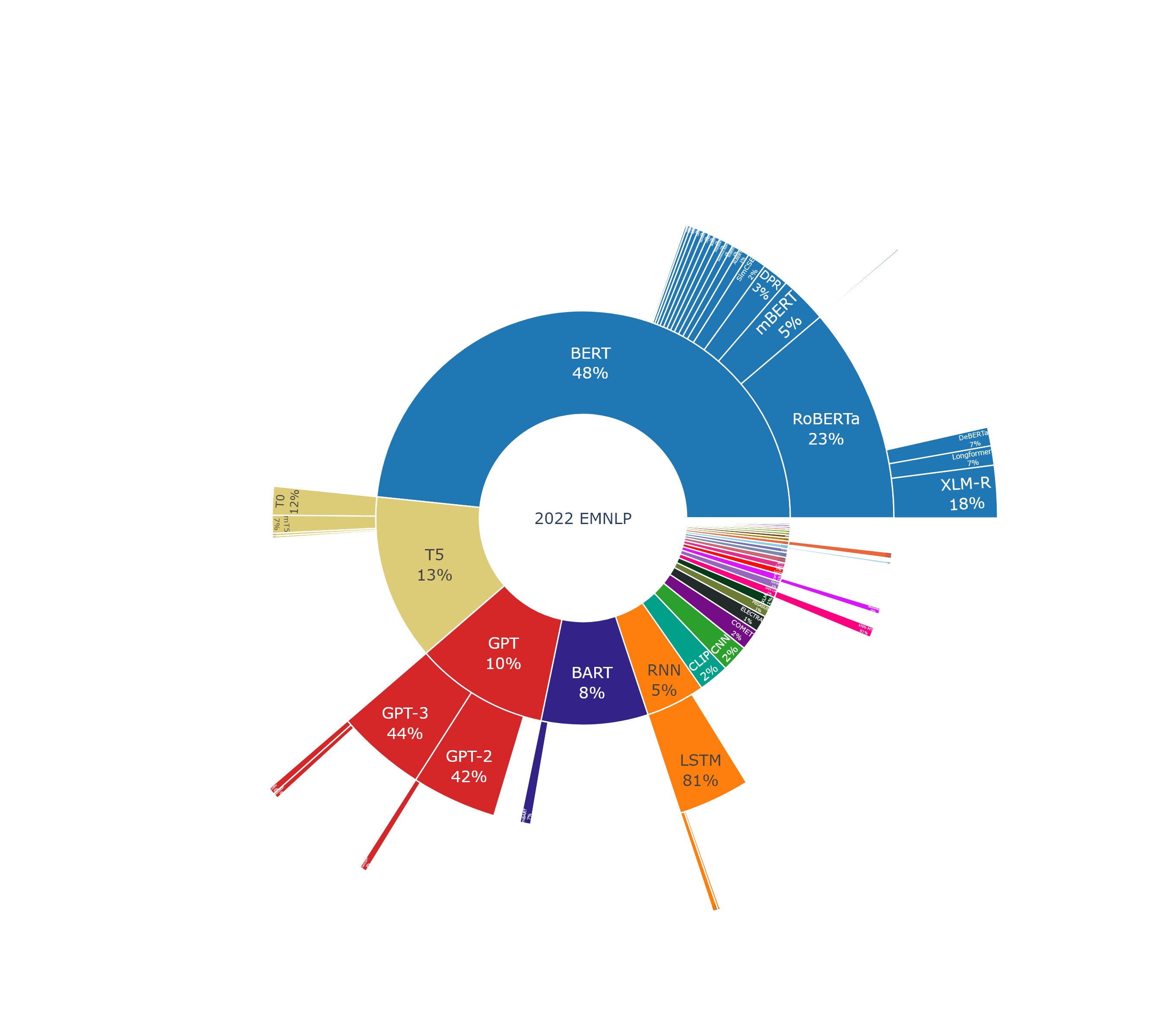}
\end{figure*}

\begin{figure*}[th!]
    \centering
    \includegraphics[width=0.88\textwidth]{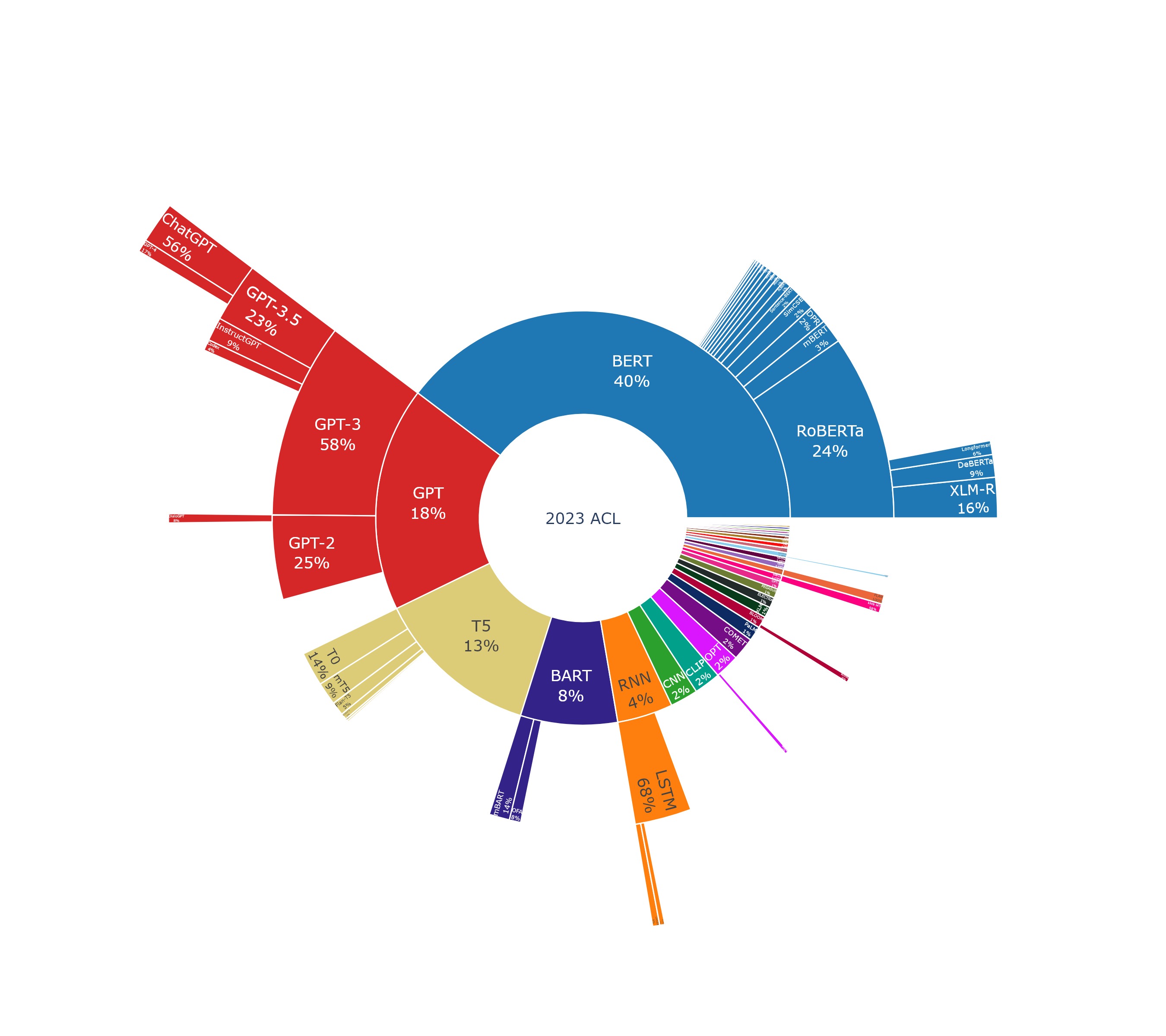}
    \vspace*{-1cm}
    \includegraphics[width=0.88\textwidth]{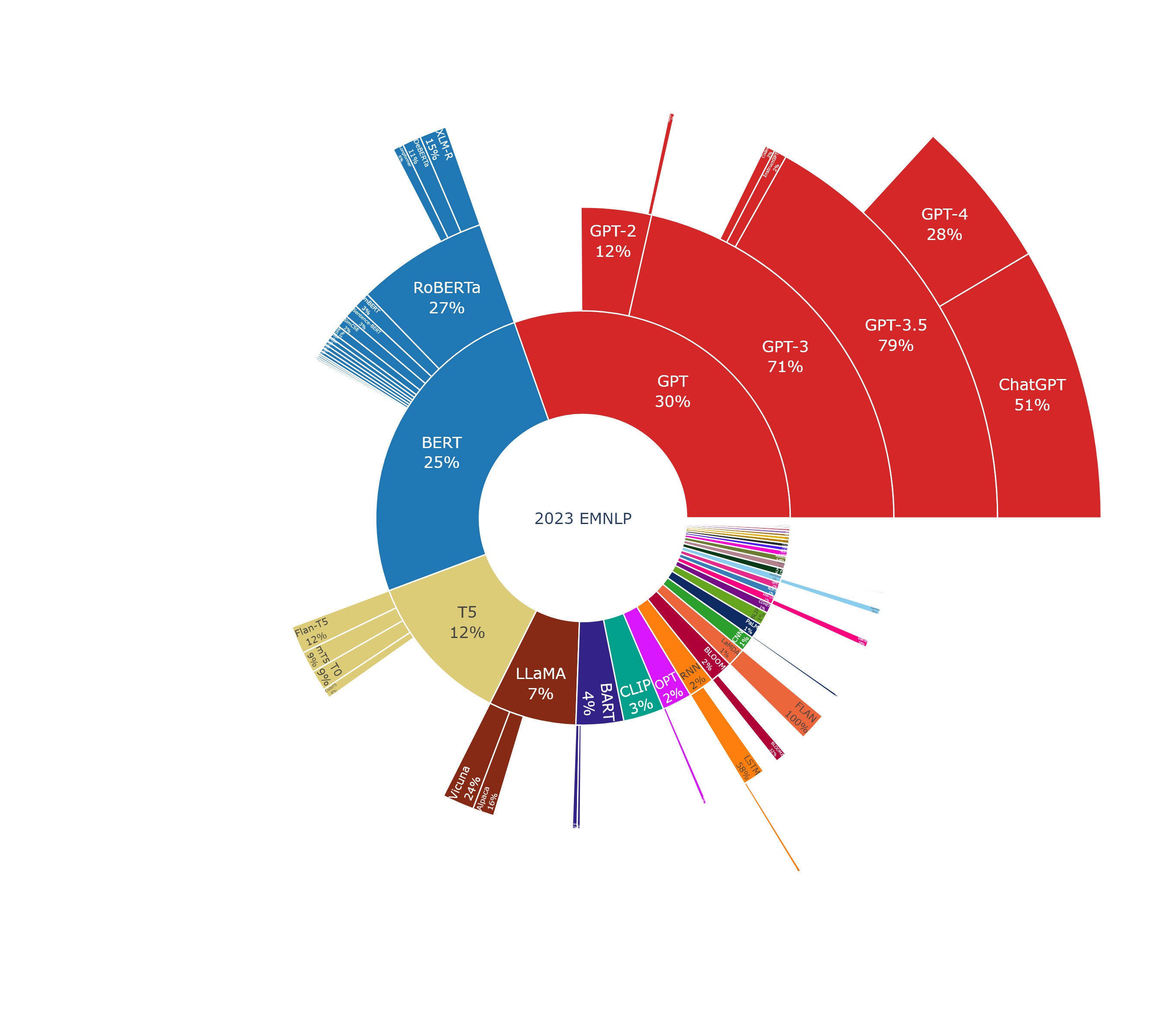}
\end{figure*}

\end{document}